\PassOptionsToPackage{dvipsnames}{xcolor}
\documentclass[11pt]{article}

\usepackage[most]{tcolorbox}
\usepackage[]{acl}
\usepackage{float}
\usepackage{placeins}
\usepackage{times}
\usepackage{latexsym}
\usepackage{multirow}
\usepackage{booktabs}
\usepackage[T1]{fontenc}

\usepackage[utf8]{inputenc}

\usepackage{adjustbox}
\usepackage{tabularx}       
\usepackage{lscape}
\usepackage{pdflscape}
\usepackage{microtype}
\usepackage{xspace}
\usepackage{arydshln}
\usepackage{hyperref}
\usepackage{amsmath}
\usepackage{inconsolata}
\usepackage{graphicx}
\usepackage{booktabs}
\usepackage{enumitem}
\usepackage{rotating}

\newcommand{\datasetname}{\texttt{RevUtil}\xspace}
\newcommand{\actionability}{\texttt{Actionability}\xspace}

\newcommand{\grounding}{\texttt{Grounding \& Specificity}\xspace}
\newcommand{\groundingshort}{\texttt{Ground. \& Spec.}\xspace}
\newcommand{\verifiability}{\texttt{Verifiability}\xspace}

\newcommand{\helpfulness}{\texttt{Helpfulness}\xspace}

\newcommand{\golddata}{\texttt{FullAgreement}\xspace}
\newcommand{\silverdata}{\texttt{MajorityAgreement}\xspace}
\newcommand{\harddata}{\texttt{LowAgreement}\xspace}

\title{The Good, the Bad and the Constructive:\\Automatically Measuring Peer Review's Utility for Authors}

\author{
Abdelrahman Sadallah,$^{1}$
Tim Baumgärtner,$^{2}$ 
Iryna Gurevych,$^{1,2}$
Ted Briscoe$^{1}$ \\
$^{1}$NLP Department, Mohamed Bin Zayed University of Artificial Intelligence \\
$^{2}$Ubiquitous Knowledge Processing Lab, Department of Computer Science \\ Hessian Center for AI (hessian.AI), TU Darmstadt \\
}

\begin{document}
\maketitle
\begin{abstract}
Providing constructive feedback to paper authors is a core component of peer review. With reviewers increasingly having less time to perform reviews, automated support systems are required to ensure high reviewing quality, thus making the feedback in reviews useful for authors. To this end, we identify four key aspects of review comments (individual points in weakness sections of reviews) that drive the utility for authors: \actionability, \grounding, \verifiability, and \helpfulness. To enable evaluation and development of models assessing review comments, we introduce the \datasetname dataset. We collect 1,430 human-labeled review comments and scale our data with 10k synthetically labeled comments for training purposes. The synthetic data additionally contains rationales, i.e., explanations for the aspect score of a review comment. Employing the \datasetname dataset, we benchmark fine-tuned models for assessing review comments on these aspects and generating rationales. Our experiments demonstrate that these fine-tuned models achieve agreement levels with humans comparable to, and in some cases exceeding, those of powerful closed models like GPT-4o. Our analysis further reveals that machine-generated reviews generally underperform human reviews on our four aspects.\footnote{Our code and data are available at \url{https://github.com/bodasadallah/RevUtil}}\looseness=-1

\end{abstract}

\section{Introduction}

\begin{figure}[ht!]
\centering
\includegraphics[width=0.9\linewidth]{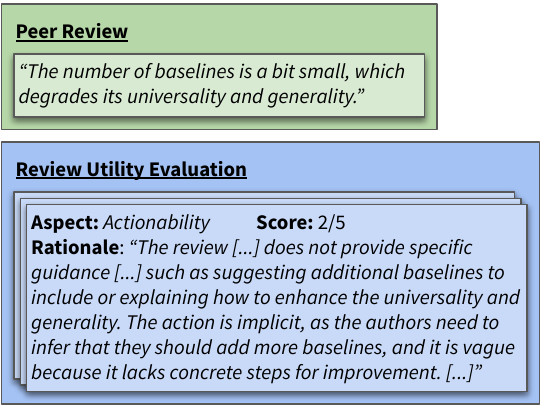}
\caption{An example from our \datasetname dataset. A single review comment, extracted from the weakness section of a peer review, is scored on a scale from 1-5 on its \actionability. In addition, a rationale for the given score is provided, explaining the score. Our dataset contains three further aspects: \grounding, \verifiability, and \helpfulness.
}
\label{fig:review-example-evaluation}
\end{figure}

Peer review is fundamental for scientific research. Besides serving as a high-quality filter, it provides explanations and constructive feedback to authors to improve their scientific contribution \citep{jefferson2002effects,ross2017open}. However, with the rapidly growing number of submissions to conferences and journals \citep{fire-over-optimization-2019,growth-rate-modern-science}, the peer review process is under pressure. Reviewers have less time to complete reviews; consequently, reviews suffer from poor quality and high variance \citep{rogers-augenstein-2020-improve,kunzli2022not}. Yet, without high-quality feedback, authors lack the guidance on how to improve their work, leading to resubmissions without substantial changes and consequent inefficiencies in the scientific process. Furthermore, unprofessional reviews can impact productivity and cause emotional distress, particularly for early-career researchers and those from underrepresented groups \citep{Silbiger2019UnprofessionalPR,hyland-jiang-2020-harsh}. Therefore, peer review would benefit from novel support systems to facilitate reviewing and increase review quality \citep{kuznetsov-et-al-2024-what-can}. 

To this end, we propose to automatically evaluate individual comments raised in the weakness section of peer reviews along four dimensions that contribute to the utility of a comment for authors. 
Utility refers to the degree to which a comment provides an \textit{actionable} critique, to a \textit{specific} part of the paper, using \textit{verifiable} arguments. In other words,  how \textit{helpful} a comment is to an author (see \S\ref{ssec:aspect_definitions} for the full definitions).
While assessing review quality automatically has many applications, ranging from better recognizing good reviewers to measuring interventions on the peer review system \citep{goldberg2024peerreviewspeerreviews}, we focus on an immediate application: providing feedback to reviewers, enabling them to revise their review, and making it more useful for authors to help them improve their paper. Fig.~\ref{fig:review-example-evaluation} shows an example, where the \actionability of a review comment is scored, and a rationale for the rating is provided.\looseness=-1

Recently, ICLR 2025 experimented with providing automatic feedback on reviews \citep{thakkar2025can}. They found that reviewers updated their review after receiving feedback, and the revised feedback was preferred by paper authors in 89\% of cases, indicating that such feedback is accepted by reviewers and effective in improving the quality and utility of reviews for authors. While this experiment is promising, research into evaluating and improving systems for automated reviewing feedback is limited by a lack of evaluation data. Besides, \citet{thakkar2025can} leveraged closed-source models to provide feedback due to their superior zero-shot capabilities. However, this raises ethical concerns as paper drafts are private, and providing them to external services can violate the ethics policies of conferences.\footnote{For example, the ACL Policy on Publication Ethics explicitly prohibits uploading submission drafts to non-privacy preserving providers \citep{acl-policy-publication-ethics}.}

\looseness=-1
Our contributions can be summarized as:\\
\textbf{1.} Introducing and defining four key aspects to assess a review comment's utility for authors, namely: \actionability, \verifiability, \grounding, and \helpfulness.
\\
\textbf{2.} Introducing the \datasetname dataset to evaluate and develop automated review assessment systems. We collect human annotations of 1,430 review comments for the introduced aspects, with each sample labeled by \underline{three} different annotators, and scale the data collection with 10k synthetically labeled review comments, using a method that obtains good agreement with human annotations.\\
\textbf{3.} Finetuning models on the collected data and show that they are equivalent or outperform closed models like GPT-4o, demonstrating that open, privacy-preserving models can provide useful review comment scoring and feedback. 
\section{Related Work}

\begin{figure*}[ht!]
\centering
\includegraphics[width=\linewidth]{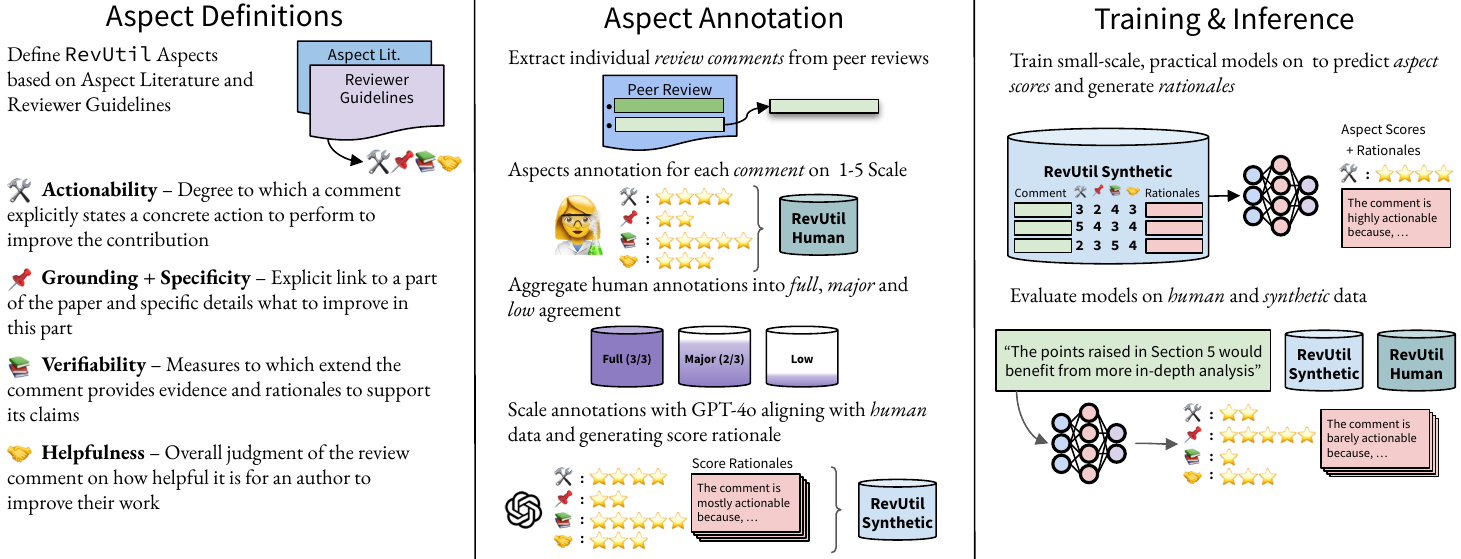}

\caption{Overview of our full annotation and modeling pipeline. We begin by defining annotation aspects, followed by collecting human-labeled data. We then generate automatic labels using existing systems, and finally fine-tune models based on the collected annotations.}
\label{fig:overview}
\end{figure*}

\subsection{Review Evaluation}
\label{sec:related-review-evaluation}
Several studies examined review quality by surveying authors, reviewers, and/or meta-reviewers on aspects such as negligence, fairness, and coverage \citep{Khosla2012AnalysisOR, stelmakh2021novice, goldberg2024peerreviewspeerreviews}. We base our aspects partially on these works, adopting and refining aspects such as substantiation (corresponding to our \verifiability), constructiveness (our \actionability), and helpfulness. However, since the data from these studies is not publicly available, it cannot be used to develop or evaluate models for automatic review assessment. Further, our data provides more fine-grained annotations by evaluating individual review comments instead of whole reviews.

\citet{purkayastha-etal-2025-lazyreview} also assess reviews, but with a different focus: They propose to detect specific ``lazy'' reviewing practices, i.e., comments that follow simple reviewing heuristics (as defined by the ACL reviewing guidelines~\citep{ARR_Reviewer_2024}). In contrast, we score review comments on a 1-5 scale for the four aspects, thereby rating the utility of the comment for the author, providing rationales for the scores, and thus guidance to the reviewer on how to improve the utility of the review for authors.\looseness=-1

Increasingly, LLMs are also used to generate peer reviews \citep{liang-monitoring-2024}. Various approaches have been put forward, testing LLM and agentic systems to generate reviews \citep{wang-etal-2020-reviewrobot,chamoun-etal-2024-automated,darcy-marg-2024,gao2024reviewer2optimizingreviewgeneration,liang-llm-feedback-generation-2024,zhu-etal-2025-deepreview}. While review generation can serve as a benchmark for LLMs, such reviews are often generic and flawed by hallucinations. A major challenge in their development is evaluating the quality of LLM-generated reviews. Comparing such reviews to human reference reviews using lexical overlap metrics is limited, since the references are diverse and the metrics do not align well with human judgments~\citep{novikova-etal-2017-need,reiter-2018-structured}. Alternatively, closed-source models such as GPT-4~\citep{liu-etal} or specialized LLM-as-a-judge models \citep{kim2024prometheus} can rate generated reviews. However, using closed-source models as judges is expensive and has limited reproducibility. For open and general LLM-as-a-judge models, our results show that they do not perform well on this task (see \S\ref{sec:results}). Conversely, our fine-tuned models are specialized in assessing an important aspect of review quality, namely the utility for authors, and can therefore be potentially used as evaluation metrics for automated review generation models.

\subsection{Review Aspects}
\label{sec:related-review-aspects}
Prior work analyses peer reviews on different aspects, including harshness \citep{verma-etal-2022-lack}, thoroughness and helpfulness \citep{severin-et-al-2022-journal-impact}, completeness \citep{yuan2022can}, substantiation \citep{guo-etal-2023-automatic}, and politeness \citep{politepper}. Closest to our work is that of \citet{severin-et-al-2022-journal-impact}. Their thoroughness aspect aims to capture the degree to which a review considered all necessary topics to judge a paper (e.g., ``\textit{Did the reviewer comment on the methods?}''). While this is interesting for a comprehensive review evaluation, it is out of scope for judging the utility of the review for authors. On the other hand, their helpfulness definition relates to our \actionability and \verifiability aspects. However, they only collect binary labels on a sentence level, employing a single annotator. In contrast, we define a detailed score rubric for these aspects on a 1-5 scale and collect three labels per sample (see \S\ref{ch:aspects}). Initially, we have also collected binary politeness data (similar to \citet{politepper}), since respectful communication is crucial for authors to be open to the provided feedback. However, we found that the vast majority of comments were appropriate in our review data and therefore decided to omit this aspect.

Albeit generated by GPT-4o, we are the first to include rationales explaining our aspect scores. Furthermore, our annotations are on the comment-level, which can consist of multiple sentences, while previous datasets provide phrase- or sentence-level annotations. A comment is the natural level of granularity as it provides the minimal, but necessary context to judge the proposed aspects, however, it relies on reviewers structuring their review accordingly.

\subsection{Human Preference Data}
Our data can also be viewed as a human preference collection, which has recently gained popularity for post-training LLMs to align them with humans \citep{ouyang-et-al-2022-instruct-gpt, bai-et-al-2022-hhrlhf}. While many human preference datasets have been collected in a pair-wise setting, where humans rank two (or more) responses, point-wise data has also shown promising results \citep{wang2024helpsteer2opensourcedatasettraining,wang-etal-2024-helpsteer}. Similarly, \datasetname contains human preferences on review comments on multiple aspects. We envision that reward models trained on our data could improve review generation using Reinforcement Learning.\looseness=-1
\section{\datasetname{}}
\label{ch:aspects}

Fig.~\ref{fig:overview} provides an overview of our proposed aspects, data collection, and model training on the review assessment task.

\subsection{Aspect Definitions}
\label{ssec:aspect_definitions}

To determine which aspects drive author utility of reviews, we study the existing aspect and review generation literature (see \S\ref{sec:related-review-evaluation} and \S\ref{sec:related-review-aspects}) and the reviewer guidelines of NLP conferences \cite{ARR_Area_Chair_2024, ARR_Reviewer_2024}. We present a brief definition for each aspect (\S\ref{app:aspects_definitions} provides detailed definitions).\looseness=-1
\smallskip

\noindent\textbf{\actionability:}
Actionability assesses the extent to which a comment offers practical guidance for the authors. It evaluates how concrete and detailed the suggestions are, and how easily they can be translated into specific actions.\looseness=-1
\smallskip

\noindent\textbf{\grounding:}
This aspect measures the extent to which a review comment is anchored to specific parts of the paper. It evaluates whether the comment clearly identifies the relevant text region and explicitly states what requires improvement.\looseness=-1
\smallskip

\noindent\textbf{\verifiability:}
This aspect evaluates whether a review comment contains a claim (i.e., a subjective opinion) and how well it is substantiated. The first step is to determine whether the comment includes any claims. If it does, we assess the extent to which the reviewer justifies or supports the claim through logical reasoning, common knowledge, or references.\looseness=-1
\smallskip

\noindent\textbf{\helpfulness:}
Helpfulness evaluates the overall usefulness of a comment to the authors, aggregating the previous three aspects. It is a subjective measure that reflects the overall value of the review comment to the authors. Therefore, this aspect integrates our three preceding, primary aspects and is a subjective rating by the annotators.\looseness=-1

\subsection{Aspect Annotation}\label{sec:raw-data-and-preprocessing}
\subsubsection{Peer Review Data}
\label{subsec:raw_data}
To collect \datasetname, we first require peer review data. We leverage the NLPeer dataset \citep{dycke-etal-2023-nlpeer} specifically using the ACL 2017 and ARR 2022 subsets. Further, we use the Reviewer2 dataset \citep{gao2024reviewer2optimizingreviewgeneration} from which we use the NeurIPS 2016-2022 and ICLR 2021-2023 data. Finally, we complement this with data obtained from OpenReview for the EMNLP 2023 and ICLR 2024-2025 conferences. As our main goal is to evaluate the utility of reviews for authors, we include only the weaknesses and discussion sections of the reviews. Other sections, such as summary and strengths, are primarily intended for meta-reviewers or chairs and hold limited value for the authors to improve their manuscript. To annotate the data effectively, we split the reviews into their individual comments. This step is crucial because assigning a single score to an entire review is challenging and prone to noise, and different review comments may vary with respect to our defined aspects (for example, some comments can be actionable, while others are not).
We apply string-matching techniques to segment the reviews into comments based on the natural formatting of the review, such as line breaks, bullet points, enumerations, etc.
Following this process, we extract 207k review comments from 60k reviews.\footnote{The review comment extraction algorithm and detailed statistics and breakdown by venue can be found in \S\ref{app:review_segmentation}.} In order to measure the accuracy of the rule-based segmentation, we manually evaluate 100 random review comments for the validity of their segmentation. We find that 94\% of the points are correctly segmented.\looseness=-1

\begin{table*}
\centering
\begin{tabular}{@{}lcccccccccc@{}} 
\toprule
 & \multicolumn{5}{c}{\textbf{\texttt{Full+Majority+Low Agreement}}} & \multicolumn{5}{c}{\textbf{\texttt{Full+Majority Agreement}}} \\
 \cmidrule(lr){2-6}
 \cmidrule(lr){7-11}
\textbf{Aspect}         & \# & \textbf{$\kappa^2$} & \textbf{$\rho$} & \textbf{$\alpha$} & F1 & \# & $\kappa^2$ & $\rho$ & $\alpha$ & F1  \\ 
\midrule
\actionability          & 1430 & 0.614 & 0.598 & 0.561 & -     & 1208 & 0.718 & 0.658 & 0.620 & -     \\
\groundingshort              & 1430 & 0.435 & 0.418 & 0.391 & -     & 1243 & 0.517 & 0.455 & 0.428 & -     \\
\verifiability          & 1430 & 0.495 & 0.479 & 0.458 & 0.189 & 1023 & 0.665 & 0.610 & 0.612 & 0.275 \\
\helpfulness            & 1430 & 0.511 & 0.478 & 0.469 & -     & 1111 & 0.664 & 0.606 & 0.602 & -     \\
\bottomrule
\end{tabular}
\caption{Inter-Annotator Agreement on the entire \datasetname dataset and the subset where all or a majority of annotators agree. We report Quadratic-Weighted Cohen's Kappa ($\kappa^2$), Spearman Correlation ($\rho$) and Krippendorf's alpha ($\alpha$). For the claim detection in the \verifiability aspect, we measure F1.}
\label{tab:iaa}
\end{table*}

\subsubsection{\datasetname{} Human}\label{ssec:human_data}
\paragraph{Data Collection}
We label the review comments on a 5-point scale (for \verifiability, we add a binary label for the claim detection task, indicating whether the comment contains a claim or not), which allows for fine-grained assessments and better reflects the utility of a review comment with respect to the different aspects.\footnote{In preliminary annotations, we investigated binary labels; however, we realized that our annotation requires a more nuanced label setup.} However, since scoring peer review quality involves a significant degree of subjectivity (as also noted by ~\citet{goldberg2024peerreviewspeerreviews}) we provide additional definitions for each score from 1-5 to provide annotators with further guidance (see \S\ref{app:aspects_definitions} for the detailed definitions). 

To obtain human labels, we recruit researchers using Prolific. We select annotators who have obtained or are currently enrolled in a PhD in CS or IT to ensure their familiarity with our peer reviews. Further, we select only native English speakers. 

We conducted three preliminary studies to identify annotators with the highest performance on the task. In total, we recruited 48 individuals across these studies. To evaluate annotator performance, we used a gold-standard dataset labeled independently by the first two authors. Any disagreements between their annotations were subsequently discussed and resolved to ensure consistency. Annotator agreement was then measured against this finalized gold-standard, and the two highest-performing annotators were selected for the main task.\looseness=-1

For our human dataset, all samples are annotated by three annotators: two recruited and the main author of this work. To maintain the quality of the annotations, we split our data into batches of 100-200 samples each, to ensure our annotators do not suffer from annotation fatigue~\citep{kern-etal-2023-annotation}.\looseness=-1

Our human-annotated dataset consists of 1,430 review comments. We create three subsets of our data: \golddata (samples for which all annotators assigned the same score), \silverdata (samples for which two annotators assigned the same score), and \harddata (samples for which the three annotators assigned different scores). We present data samples in \S\ref{app:data-samples}, and report statistics and label distribution in \S\ref{app:manual_data_statistics}.\looseness=-1

\paragraph{Inter-Annotator Agreement}
To evaluate the difficulty of the annotation task and understand the level of subjectivity involved, we compute the inter-annotator agreement (IAA). Following previous work with a similar annotation scheme \citep{wang2024helpsteer2opensourcedatasettraining}, we use quadratic-weighted Kappa ($\kappa^2$) \citep{cohen1968weighted}, an extension of Cohen's Kappa that accounts for the magnitude of disagreements, making it suitable for ordinal data. We also report Spearman correlation ($\rho$) and Krippendorff's alpha ($\alpha$). For the claim detection task in the \verifiability aspect, we report F1, and exclude the samples with ``No Claim'' labels from the subsequent agreement calculations. For the pair-wise metrics ($\kappa^2, \rho$, F1), we compute the average over the three annotator pairings.\looseness=-1

Table~\ref{tab:iaa} presents the agreement across the four annotated aspects, for the entire \datasetname Human data, and for the subset where a majority among annotators is reached. Focusing on $\kappa^2$, we observe moderate agreement levels across all aspects. The highest $\kappa^2$ score (0.614) occurs for \actionability, suggesting that annotators were relatively consistent when determining whether the comments included actionable comments. \grounding shows a lower $\kappa^2$ score (0.435), indicating greater subjectivity in this aspect. Furthermore, the low value of the F1 comes from divergent annotations from one annotator (see Table~\ref{tab:human_annotators_agreement_pairwise} for the pair-wise agreements). When excluding the annotations for which no agreement can be reached, we retain 72\%-84\% of our original data. On this subset, we obtain a substantial agreement ($\kappa^2>0.6$) for all aspects, except \grounding, which improves, but remains in the moderate agreement range.

\subsubsection{\datasetname{} Synthetic}
\label{ssec:synthetic_data}
As collecting human annotations is costly, and the capacity of our manual data is not enough to perform model fine-tuning, we employ GPT-4o\footnote{Specifically, \texttt{GPT-4o-2024-11-20}.} to generate more synthetic annotations.

\begin{table}
    \centering

    \resizebox{\linewidth}{!}{ 
    
\begin{tabular}{@{}lccccc@{}} 
\toprule
\textbf{Aspect}       & \# & $\kappa^2$ & $\rho$ & $\alpha$ & F1 \\ 
\midrule
\actionability         & 1208 & 0.544 & 0.628 & 0.450 & -      \\
\groundingshort             & 1243 & 0.517 & 0.546 & 0.411 & -      \\
\verifiability         & 1023 & 0.368 & 0.564 & 0.151 & 0.533  \\
\helpfulness           & 1111 & 0.544 & 0.608 & 0.456 & -      \\
\bottomrule   
\end{tabular}

    }
    
    \caption{Agreement for aspects between the synthetically generated labels by GPT-4o, and the human \golddata+ \silverdata data. Each aspect has a different total number of examples (\#), as the number of samples where humans reach a majority agreement differs for each aspect. 
    }
    \label{tab:synthetic_data_ageement_gold_silver_agreement}
\end{table}

We provide the aspect definition and in-context examples \citep{gpt3} in the prompt. For each score, we manually label 10 samples and write rationales for labeling them respectively (for the claim detection, we craft an additional 30 samples, 15 containing a claim, 15 without a claim). For each sample to annotate, we sample 5 examples per score and add them as in-context examples to the prompt, resulting in 25 in-context examples (10 for claim detection, as there are only two labels).\footnote{We also tried to incorporate Chain-of-Thought reasoning~\cite{wei2023chainofthoughtpromptingelicitsreasoning} in our prompt, but found that providing rationales along with the examples is more effective. The same conclusion has also been reached in similar work~\citep{calderon2025alternativeannotatortestllmasajudge}.} Subsequently, each aspect is annotated in a separate request. For \verifiability, we perform the annotation in two steps, first, we detect whether the review comment contains any claims. Then, we run the scoring for those review comments that have previously been determined to contain a claim. In total, we generate 10,000 labeled review comments for each aspect. The exact prompt template can be found in \S\ref{app:prompts}, and \S\ref{app:synthetic_data_statistics} reports the statistics of the synthetic data.\looseness=-1

To assess the quality of our synthetic data, we also obtain labels for the review comments previously labeled by humans. Table~\ref{tab:synthetic_data_ageement_gold_silver_agreement} reports the agreement statistics using the \golddata and \silverdata samples. Across all aspects, the $\kappa^2$ scores range from 0.368 to 0.544, indicating fair to moderate agreement between synthetic and human labels.\looseness=-1

\begin{table*}[t]
\centering
\resizebox{1.0\textwidth}{!}{

\begin{tabular}{@{}lcccccccccccccc@{}} 
\toprule
\multicolumn{3}{c}{}                                         & \multicolumn{2}{c}{\textbf{\actionability}}                 & \multicolumn{2}{c}{\textbf{\groundingshort}}                     & \multicolumn{4}{c}{\textbf{\verifiability}}                                                             & \multicolumn{2}{c}{\textbf{\helpfulness}} & \multicolumn{2}{c}{\textbf{Average}}                 \\
\multicolumn{3}{c}{}                                         & \textbf{Synth.}       & \multicolumn{1}{c}{\textbf{Human}} & \textbf{\textbf{Synth.}} & \multicolumn{1}{c}{\textbf{\textbf{Human}}} & \multicolumn{2}{c}{\textbf{\textbf{Synth.}}} & \multicolumn{2}{c}{\textbf{\textbf{Human}}}             & \textbf{\textbf{Synth.}} & \textbf{\textbf{Human}} & \textbf{\textbf{Synth.}} & \textbf{\textbf{Human}}       \\ 
\cmidrule(lr){4-5}
\cmidrule(lr){6-7}
\cmidrule(lr){8-11}
\cmidrule(lr){12-13}
\cmidrule(lr){14-15}
\textbf{Model} & \textbf{Mode}        & \textbf{Type}        & $\kappa^2$ & $\kappa^2_{avg}$       & $\kappa^2$    & $\kappa^2_{avg}$                & $\kappa^2$ & F1          & $\kappa^2_{avg}$ & F1$_{avg}$ & $\kappa^2$ & $\kappa^2_{avg}$ & $\kappa^2$ & $\kappa^2_{avg}$  \\ 
\midrule
\multicolumn{15}{c}{\textbf{Zero-Shot}}                 \\ 
\hdashline

GPT-4o         & S   & Inst. & \textbf{0.696}     & \textbf{0.397} & \textbf{0.591}     & \textbf{0.517} & 0.433     & \textbf{0.598}     & 0.150 & \textbf{0.314} & \textbf{0.582}     & \textbf{0.396} & \textbf{0.576} & \textbf{0.365} \\
GPT-4o         & S+R & Inst. & \underline{0.493}     & 0.279 & \underline{0.434}     & \underline{0.464} & \textbf{0.479}     & \underline{0.304}     & 0.184 & 0.132 & \underline{0.530}     & \underline{0.291} & \underline{0.484} & \underline{0.305} \\
Prometheus-2-7B & S   & Inst. & 0.127 & 0.075 & 0.109 & 0.115 & 0.147 & 0.008 & 0.174 & 0.027 & 0.318 & 0.199 & 0.175 & 0.141 \\
Prometheus-2-7B & S+R & Inst. & 0.309 & 0.172 & 0.134 & 0.084 & 0.239 & 0.134 & 0.129 & 0.127 & 0.204 & 0.148 & 0.222 & 0.133 \\
Selene-1-Mini-8B & S   & Inst. & 0.298 & 0.187 & 0.215 & 0.197 & 0.154 & 0.077 & \textbf{0.281} & 0.017 & 0.372 & 0.291 & 0.260 & 0.239 \\
Selene-1-Mini-8B & S+R & Inst. & 0.190  & 0.100   & 0.388 & 0.328 & \underline{0.444} & 0.093 & 0.193 & 0.041 & 0.438 & 0.204 & 0.365 & 0.206 \\
Llama-3.1-8B         & S   & Inst. & 0.057 & 0.045 & 0.155 & 0.153 & 0.095 & 0     & 0.148 & 0     & 0.155 & 0.142 & 0.116 & 0.122 \\
Llama-3.1-8B         & S+R & Inst. & 0.155 & 0.082 & 0.099 & 0.058 & 0.042 & 0.048 & 0.085 & 0.024 & 0.096 & 0.066 & 0.098 & 0.073 \\
Llama-3.1-IT-8B       & S   & Chat  & 0.333 & \underline{0.311} & 0.196 & 0.271 & 0.046 & 0.159 & 0.177 & 0.041 & 0.220  & 0.247 & 0.199 & 0.252 \\
Llama-3.1-IT-8B       & S+R & Chat  & 0.103 & 0.060  & 0.221 & 0.287 & 0.420  & 0.005 & 0.237 & 0.007 & 0.426 & 0.270  & 0.293 & 0.214 \\
DeepSeek-R1-7B       & S   & Chat  & 0.301 & 0.192 & 0.361 & 0.261 & 0.337 & 0.114 & \underline{0.250}  & 0.040  & 0.387 & 0.269 & 0.347 & 0.243 \\
DeepSeek-R1-7B       & S+R & Chat  & 0.255 & 0.156 & 0.319 & 0.224 & 0.284 & 0.110  & 0.214 & 0.054 & 0.323 & 0.249 & 0.295 & 0.211 \\
SciLitLLM-7B         & S   & Chat  & 0.087 & 0.052 & 0.012 & 0.024 & 0.069 & 0     & 0.044 & 0.004 & 0.321 & 0.139 & 0.122 & 0.065 \\
SciLitLLM-7B         & S+R & Chat  & 0.289 & 0.157 & 0.194 & 0.093 & 0.036 & 0.199 & 0.038 & \underline{0.150}  & 0.206 & 0.123 & 0.181 & 0.103 \\
\midrule
\multicolumn{15}{c}{\textbf{Fine-Tuning}}                 \\ 
\hdashline
Llama-3.2-IT-3B & S                      & Inst.                  & 0.684 & 
0.464 & 0.657 & 0.479&  0.648 & \underline{0.611} & \textbf{0.356}   & 0.263 &0.660& 0.453 & 0.662 & 0.438
\\
Llama-3.1-8B & S     & Inst. & 0.724 & \underline{0.498} & 0.636 & 0.473 & 0.640 & 0.599 & 0.333 & 0.265 & \textbf{0.712} & 0.469 & 0.678 & 0.443 \\
Llama-3.1-8B & S+R & Inst. & 0.705 & 0.478 & 0.674 & 0.482 & \textbf{0.669} & 0.466 & 0.329 & 0.258 & \underline{0.683} & \textbf{0.474} & \underline{0.683} & 0.441 \\ 
Llama-3.1-IT-8B    & S   & Inst. & 0.709 & \textbf{0.510} & \underline{0.692} & \textbf{0.526} & 0.642 & \textbf{0.644} & \underline{0.350} & \underline{0.266} & 0.681 & \textbf{0.474} & 0.681 & \textbf{0.465} \\
Llama-3.1-IT-8B    & S+R & Inst. & \underline{0.726} & 0.492 & \textbf{0.699} & \underline{0.517} & \underline{0.659} & 0.439 & 0.319 & 0.226 & 0.663 & 0.455 & \textbf{0.687} & \underline{0.446} \\
DeepSeek-R1-7B       & S   & Inst. & 0.532 & 0.381 & 0.421 & 0.381 & 0.581 & 0.475 & 0.300 & 0.243 & 0.539 & 0.381 & 0.518 & 0.361 \\
DeepSeek-R1-7B       & S+R & Inst. & \textbf{0.733} & 0.453 & 0.691 & 0.497 & 0.613 & 0.512 & 0.278 & 0.254 & 0.652 & 0.393 & 0.672 & 0.405 \\
SciLitLLM-7B         & S   & Inst. & 0.653 & 0.454 & 0.548 & 0.433 & 0.581 & 0.508 & 0.316 & \textbf{0.274} & 0.626 & 0.418 & 0.602 & 0.405 \\
SciLitLLM-7B         & S+R & Inst. & 0.664 & 0.472 & 0.610  & 0.469 & 0.546 & 0.353 & 0.275 & 0.196 & 0.631 & 0.407 & 0.613 & 0.406 \\

\bottomrule
\end{tabular}
}
\caption{Model performance on predicted aspect scores across the test split of the synthetic (Synth.), and the human data on \datasetname. We evaluate models in a zero-shot regime using their native templates and after fine-tuning on the synthetic training data in an instruction-based setting. We evaluate two generation modes: to only generate a score (S), and to generate a rationale before predicting the score (S+R). 
}
\label{tab:main_results}
\end{table*}

\section{Experiments}
\label{sec:finetuning}
In the previous section, we introduced the \datasetname data. While GPT-4o (or similar closed-source models) can be leveraged to perform review comment assessment, it is unsuitable in a practical setting for several reasons: Manuscript drafts are confidential and should therefore not be sent to potentially non-privacy-preserving services. This violates, for example, the ARR guidelines. Moreover, using third-party services can be expensive, especially given high volumes of submissions at conferences.

Consequently, we experiment with small-scale, open-weight models. As a baseline, we first use them in a zero-shot setting, before fine-tuning them on our data. For fine-tuning, we utilize LoRA ~\cite{hu2021loralowrankadaptationlarge} to maximize the utilization of our computational resources. Previous research demonstrated that LoRA achieves comparable performance to full fine-tuning, while requiring significantly less time and compute \cite{dettmers2023qloraefficientfinetuningquantized,ghosh2024closerlooklimitationsinstruction}.
While we labeled aspects individually for our synthetic data to maximize quality, we opt to train a single model for all aspects, requiring only a single call to predict all four aspects during inference. Fine-tuning LLMs requires selecting an appropriate prompt template. Most fine-tuned models incorporate the chat templates used during their fine-tuning process. However, these chat templates are primarily designed for multi-turn conversations, whereas our task involves single-turn response generation. Consequently, we adopted the Alpaca-style prompt template~\cite{peng2023instructiontuninggpt4}. The specific prompts used in our experiments are detailed in \S\ref{app:prompts}.
We observed that providing GPT-4o with rationales in the in-context examples improved the quality of generated synthetic labels, which aligns with previous findings~\citep{ye2023flask,zheng2023judging}. To investigate whether this approach also enhances the fine-tuning process, we design two experiments: one where the model is trained to generate only scores, and another where it is instructed to first generate a rationale before producing the score. We split the synthetic data randomly, using 90\% for training and 10\% for testing. We conducted a preliminary evaluation with SciLitLLM to determine a reasonable learning rate and LoRA rank. Finally, we evaluate on the synthetic test set and human data. We use $\kappa^2$ as our main evaluation metric. On the human data, we average the $\kappa^2$ scores over the annotators, to which we refer to as $\kappa^2_{avg}$.\looseness=-1

We select models based on two criteria: moderate size ($\leq$8 billion parameters), and generalist models vs science-specific models, to evaluate the impact of specialization on task performance. Based on these criteria, we selected the following models: Llama-3.2-3B-Instruct, Llama-3.1-8B, Llama-3.1-8B-Instruct~\cite{grattafiori2024llama3herdmodels}, DeepSeek-R1-Distill-Qwen-7B, DeepSeek-R1-Distill-Llama-70B~\cite{deepseekai2025deepseekr1incentivizingreasoningcapability}, Gemma-3-27b-IT~\citep{gemma-3-technical-report} representing generalist models, and
SciTülu-7B~\cite{wadden2024sciriffresourceenhancelanguage}, SciLitLLM~\cite{li2025scilitllmadaptllmsscientific}, DeepReview~\cite{zhu-etal-2025-deepreview}, representing models specialized to scientific tasks. We further evaluate fine-tuned LLM-Judges in the zero-shot setting, specifically Prometheus-2~\citep{kim2024prometheus}, Flow-Judge-v0.1 \citep{flow-judge}, and Selene-1-Mini-Llama-3.1-8B~\citep{alexandru2025atlaseleneminigeneral}.

\section{Results}\label{sec:results}

Table~\ref{tab:main_results} reports the results of our zero-shot and fine-tuning experiments for the best-performing models. Full results are in \S\ref{app:full_results}.\footnote{Some of the models in the zero-shot evaluation failed to produce valid outputs (outputs for all aspects, and with valid scores) for all samples. For example, for the Prometheus-2 model under the \textit{Score+Rationale} setting, we could only extract outputs for 71\% of the samples.}\\
\textbf{Synthetic vs Human} Overall, we observe that agreement scores on the human-annotated dataset are generally lower than on the synthetic test set. This is expected, as the human agreement scores represent the average agreement against the three annotators, including \harddata cases where all three annotators provided different scores.\\
\textbf{Zero-Shot} In the zero-shot setting, GPT-4o demonstrates the strongest overall performance across most evaluation aspects, except for \verifiability. Unsurprisingly, the small, open models underperform GPT-4o without fine-tuning. However, even the tested LLM judges that are fine-tuned to rate texts according to provided score definitions, severely lag behind GPT-4o.\\
\textbf{Fine-Tuning} Fine-tuning greatly boosts performance compared to zero-shot, with Llama-3.1-IT-8B achieving the highest agreement across all aspects (e.g., \actionability rising from 0.103 to 0.726). Training models to generate rationales yields only limited, inconsistent gains, and science-specialized models (ScitLitLLM, DeepReviewer, SciTülu) underperform Llama-3.1-IT-8B.

\begin{table}
\centering
\resizebox{1\linewidth}{!}{
\begin{tabular}{@{}lcccccc@{}} 
\toprule
\multicolumn{1}{c}{}  & \multicolumn{3}{c}{\textbf{GPT-4o}}                      & \multicolumn{3}{c}{\textbf{Llama-3.1-IT-8B}}         \\ 
\cmidrule(lr){2-4}
\cmidrule(lr){5-7}
\textbf{Aspect}       & $\kappa^2$ & $\rho$ & F1 & $\kappa^2$ & $\rho$ & F1 \\ 
\midrule
\actionability         & 0.544                 & 0.628               & -           & 0.550                 & 0.604               & -            \\
\groundingshort & 0.517                 & 0.546               & -           & 0.563                 & 0.545               & -            \\
\verifiability         & 0.368                 & 0.564               & 0.533       & 0.389                 & 0.621               & 0.476        \\
\helpfulness           & 0.544                 & 0.608               & -           & 0.554                 & 0.617               & -            \\
\bottomrule
\end{tabular}
}
\caption{Comparison of agreement scores between labels generated by GPT-4o (i.e., our synthetic data), and labels generated using the best fine-tuned model Llama-3.1-IT-8B on the combined \golddata and \silverdata data.
}
\label{tab:chatgpt_vs_best_model}

\end{table}

To motivate the benefits of fine-tuning, we compare Llama-3.1-IT-8B with GPT-4o on aspect prediction (Table~\ref{tab:chatgpt_vs_best_model}). While GPT-4o uses separate predictions with in-context examples, Llama-3.1-IT-8B predicts all aspects simultaneously in \textit{score only} mode. The fine-tuned model outperforms GPT-4o in $\kappa^2$, but underperforms in claim detection F1, likely due to GPT-4o’s two-step approach for \verifiability compared to the single-step setup.\looseness=-1

\paragraph{Rationale Evaluation}

We evaluated GPT-4o’s synthetic rationales by sampling 25 test examples per review aspect (100 in total) and rating them on Relevance and Correctness using a 5-point Likert scale. We define Relevance as the extent to which a rationale appropriately addresses the review comment and refers to its key points. Correctness measures whether the rationale follows the established scoring guidelines and provides a valid justification for the score according to the annotation guidelines.

\begin{table}
\centering
\resizebox{1.0\linewidth}{!}{

\begin{tabular}{lllll} 
\toprule
\textbf{Metric} & \textbf{Action.} & \textbf{Ground.} & \textbf{Verify.} & \textbf{Help.}  \\ 
\midrule
\multicolumn{5}{c}{\textbf{GPT-4o}}                                                                 \\ 
\hdashline
Relevance       & 4.64             & 4.52             & 4.28             & 4.92           \\
Correctness     & 4.16             & 4.32             & 3.64             & 4.40            \\ \midrule
\multicolumn{5}{c}{\textbf{Llama-3.1-IT-8B}}                                                  \\ 
\hdashline
Relevance       & 4.63             & 4.33             & 4.42             & 4.75            \\
Correctness     & 4.04             & 4.25             & 3.92             & 4.46            \\
\bottomrule
\end{tabular}
}

\caption{Human evaluation scores of the rationales of 25 test samples from the synthetic \datasetname data.
}
\label{tab:rationale_eavaluation}
\end{table}

Table~\ref{tab:rationale_eavaluation} shows that the average ratings for both Relevance and Correctness are high, indicating that the synthetic rationales are of strong overall quality. Based on this positive outcome, we use these synthetic rationales (for the full synthetic test set) as references to automatically evaluate the rationales generated by different models. Table~\ref{tab:rotated_rationale_scores} presents the results of this automatic evaluation using Rouge-L~\citep{rouge} and BERTScore~\citep{bertscore}. We report scores separately for correctly predicted aspect scores (subscript $C$) and incorrectly predicted ones ($W$), considering a prediction to be correct if it is within $\pm 1$ of the reference score.\looseness=-1

Across all four aspects, fine-tuned models outperform zero-shot ones in both lexical and semantic similarity. GPT-4o achieves the highest similarities with the reference rationales in the zero-shot setting. However, this might also be due to the rationales coming from the same underlying model. Among the fine-tuned models, Llama-3.1-IT-8B  achieves the highest scores across all metrics, followed closely by SciLitLLM, which aligns with the findings observed in Table~\ref{tab:main_results}. Notably, the metrics are higher when the associated aspect prediction is correct, reinforcing the link between accurate scoring and high-quality explanations. To further verify the quality of the rationales generated by the fine-tune models, we manually evaluate the generated rationales by our best model on the same 25 test samples. Results in Table~\ref{tab:rotated_rationale_scores} show that the rationales are on par with the GPT-4o rationales.

\begin{table}
\centering
\resizebox{\linewidth}{!}{%
\begin{tabular}{lccccc} 
\toprule
 & \multicolumn{3}{c}{\textbf{Zero-Shot}} &  \multicolumn{2}{c}{\textbf{Fine-Tuning}} \\ 
 \cmidrule(lr){2-4}
 \cmidrule(lr){5-6}
\textbf{Metric} & \textbf{GPT-4o} & \textbf{Llama-3.1-IT} & \textbf{SciLitLLM} & \textbf{Llama-3.1-IT} & \textbf{SciLitLLM}  \\ 
\midrule
\multicolumn{6}{c}{\textbf{\actionability}}                                                                                      \\ 
\hdashline
$R_C$           & 0.355                & 0.219                   & 0.222                   & 0.503                  & 0.487                    \\
$R_W$           & 0.328                & 0.210                   & 0.213                   & 0.418                  & 0.409                    \\
$BS_C$          & 0.720                & 0.634                   & 0.618                   & 0.785                  & 0.777                    \\
$BS_W$          & 0.694                & 0.610                   & 0.602                   & 0.736                  & 0.733                    \\ 
\midrule
\multicolumn{6}{c}{\textbf{\grounding}}                                                                                          \\ 
\hdashline
$R_C$           & 0.362                & 0.247                   & 0.231                   & 0.556                  & 0.546                    \\
$R_W$           & 0.278                & 0.222                   & 0.215                   & 0.389                  & 0.387                    \\
$BS_C$          & 0.703                & 0.630                   & 0.607                   & 0.799                  & 0.794                    \\
$BS_W$          & 0.672                & 0.626                   & 0.579                   & 0.736                  & 0.730                    \\ 
\midrule
\multicolumn{6}{c}{\textbf{\verifiability}}                                                                                      \\ 
\hdashline
$R_C$           & 0.319                & 0.216                   & 0.213                   & 0.513                   & 0.499                    \\
$R_W$           & 0.284                & 0.214                   & 0.197                   & 0.366                   & 0.367                    \\
$BS_C$          & 0.685                & 0.612                   & 0.592                   & 0.773                   & 0.769                    \\
$BS_W$          & 0.652                & 0.589                   & 0.572                   & 0.691                   & 0.692                    \\ 
\midrule
\multicolumn{6}{c}{\textbf{\helpfulness}}                                                                                        \\ 
\hdashline
$R_C$           & 0.264                & 0.189                   & 0.203                   & 0.475                   & 0.454                    \\
$R_W$           & 0.221                & 0.172                   & 0.192                   & 0.381                   & 0.357                    \\
$BS_C$          & 0.674                & 0.608                   & 0.609                   & 0.777                   & 0.769                    \\
$BS_W$          & 0.633                & 0.590                   & 0.588                   & 0.736                   & 0.726                    \\
\bottomrule
\end{tabular}
}
\caption{
Rouge-L ($R$) and BERTScore ($BS$) metrics for the generated rationales compared with the rationales in the \datasetname Synthetic test set, split by Correct (C), and incorrect (W) examples.
}
\label{tab:rotated_rationale_scores}
\end{table}
\section{Analysis}

\subsection{Error Analysis}
To examine the limitations of both human and synthetic data, we sample 10 comments per aspect from the human and synthetic data, yielding 80 examples. We analyze cases of disagreement. \looseness=-1

\paragraph{Annotator Disagreement}
We focus on 40 samples from the \harddata subset, where all annotators assigned different labels. Three recurring patterns emerge: (1) scores span three consecutive values (e.g., 1-3); (2) two low vs. one high score; and (3) two high vs. one low score. Disagreements often resulted from subtle differences in interpretation:\looseness=-1
\begin{itemize}[label={}, left=-5pt, noitemsep,topsep=0pt]
\looseness=-1
    \item \textbf{\actionability}: Annotators disagree on the level of detail required for a comment to be deemed actionable. For instance, ``The architecture used for the experiments is not clearly explained'' is given a high score by two annotators, and a low one by the third.
    \looseness=-1
    \item \textbf{\grounding}: Since this aspect exhibited the lowest IAA ($\kappa^2 = 0.435$, cf. Table~\ref{tab:iaa}), we perform a more detailed analysis of 187 samples without a majority agreement. For this, we analyze the pair-wise disagreements between each annotator. Quantitatively, we find that 40.3\% only differ by 1, 36\% differ by 2, 17.5\% differ by 3, and 6.2\% differ by 4. This shows that even in the relatively small set of low agreement (13\% of all data), a general trend of agreement persists. Finally, our qualitative analysis of these samples suggests that the primary reason for these disagreements stems from the subjective interpretation of implicit references to specific parts of the paper. While implicit references are allowed, annotators vary in judging whether vague references (``The method is only tested on two datasets'') are identifiable to authors.
    
    \item \textbf{\verifiability}: Disagreement arises when annotators interpret borderline comments as either facts or claims.
    \item \textbf{\helpfulness}: Most cases (80\%) follow the three-consecutive-score pattern.
\end{itemize}

\looseness=-1
\paragraph{Human vs. Synthetic Disagreement} We examine 40 cases where synthetic scores deviate from the majority human rating by more than 2 points and make the following observations:
\begin{itemize}[label={}, left=-5pt, noitemsep,topsep=0pt]
\item \textbf{\actionability}: In 90\% of cases, the model assigns lower scores than humans. It tends to treat reviewer questions as vague and non-actionable, despite our definitions stating that questions are actionable if they direct authors to address or clarify specific points in the draft.

\item \textbf{\grounding}: Similarly, the model frequently underrates comments referencing the overall paper (e.g., writing, formatting), failing to recognize them as grounded, despite their implicit reference to identifiable paper-wide issues.

\item \textbf{\verifiability}: Disagreement often stems from claim detection. The model either overlooks claims or misinterprets neutral statements as claims.\looseness=-1

\item \textbf{\helpfulness}: Most errors involve underestimation, similar to the patterns found in \actionability, and \grounding.
\end{itemize}

\subsection{Utility of Human vs. LLM Reviews}
\begin{table}[h]
\centering
\resizebox{\linewidth}{!}{  

\begin{tabular}{lccc} 
\toprule
\textbf{Aspect} & \textbf{Human}                   & \textbf{GPT-4}                   & $p$                        \\ 
\midrule
\multicolumn{4}{c}{Llama-3.1-IT-8B}                                                                                \\ 
\hdashline
\actionability  & $\mathbf{2.34 \pm 1.36}$             & $1.89 \pm 0.91$                      & 0.023                     \\
\groundingshort & $\mathbf{2.92 \pm 1.53}$             & $2.43 \pm 0.88$                      & 0.017                     \\
\verifiability  & $\mathbf{2.14 \pm 1.24}$             & $1.66 \pm 0.84$                      & 0.009                     \\
\helpfulness    & $2.81 \pm 1.06$                      & $\mathbf{2.96 \pm 0.72}$             & 0.335                     \\ 
\hline
\multicolumn{4}{c}{Human Evaluation}                                                                               \\ 
\hdashline
\actionability  & $\mathbf{3.15 \pm 1.72}$  & $2.91 \pm 1.66$  & 0.440   \\
\groundingshort & $\mathbf{3.28 \pm 1.37}$  & $2.91 \pm 1.26$  & 0.123  \\
\verifiability  & $\mathbf{3.30 \pm 1.57}$  & $2.94 \pm 1.37$  & 0.236  \\
\helpfulness    & $\mathbf{3.16 \pm 1.44}$ & $2.98 \pm 1.25$ & 0.448  \\
\bottomrule
\end{tabular}
}

\caption{Average aspect scores (and standard deviation) rated by our fine-tuned Llama-3.1-IT-8B model, and by manual evaluation, for Human-, and GPT-4 generated reviews on 20 papers from \citet{du-etal-2024-llms}.}
\label{tab:human_vs_llm_quality}
\end{table}

To investigate differences in review utility between human-written and LLM-generated reviews, we analyze a subset of 20 papers from~\citet{du-etal-2024-llms}. For our comparison, we select one human-written and one GPT-4-generated review per paper from their dataset. Each review is manually segmented into comments, resulting in 95 review comments from human reviews and 47 from LLM-generated reviews with average lengths of 42.6 and 26.4 words, respectively.

We assess the utility of each comment using our best-performing fine-tuned model, Llama-3.1-IT-8B, and also perform a human evaluation.\footnote{Human evaluation conducted by the first author.} Table~\ref{tab:human_vs_llm_quality} presents the average scores across our four aspects. The results show that human-written reviews outperform LLM-generated ones across 3 out of 4 aspects with statistically significant differences measured by Welch's t-test~\citep{welch}. On the \helpfulness aspect, where GPT-4 scores are slightly higher, the difference in the reviews is not significant. We conjecture that the increased performance on this aspect stems from the training regime of GPT-4, where Helpfulness has likely been incorporated during post-training, as has been reported for its predecessor InstructGPT \citep{ouyang-et-al-2022-instruct-gpt}. The same trends repeat with the human evaluation, where human-written reviews are rated with higher utility than the LLM-generated ones, albeit not being statistically significant.

\section{Conclusion}
In this work, we focused on assessing the utility of peer reviews for authors. We defined the task by identifying four aspects with clear definitions and score categories for each. We use these definitions to collect \datasetname, a dataset of 1,430 human, and 10k synthetically annotated review comments for each of our four aspects. We fine-tuned models on our newly collected data and showed that they can outperform zero-shot, closed-source models. We envision our data and models to be used to open new avenues for building automated systems that can provide real-time feedback on review quality, fostering more constructive and beneficial peer reviews.

\section*{Limitations}

\paragraph{Limited Domains and Languages}
\datasetname exclusively contains peer reviews from the NLP and ML community. Our evaluation is carried out on reviews in these domains, and whether our models generalize to other domains in Computer Science or to other scientific domains is unclear. We leveraged synthetic data generation, which requires development data to ensure quality. In this way, training data could also be collected for other scientific domains. While our developed aspects are derived from NLP reviewing guidelines, we believe they are core aspects for peer review, and our framework is transferable to other domains. Specifically, the reviewing guidelines of other conferences and journals also mention our aspects. For example, Nature \citep{nature-reviewing-guidelines} mentions in their guidelines policies relating to our \verifiability aspect (``\textit{All statements should be justified and argued in detail, naming facts and citing supporting references}''), the CVF conferences advise reviewers to ``\textit{Be generous about giving the authors new ideas for how they can improve their work}'' \citep{cvpr-reviewing-guidelines}, to ``\textit{be specific}'' and ``\textit{Give Feedback to Improve Submissions}'' \citep{iccv-reviewing-guidelines} relating to our \actionability and \grounding, or PLOS mentions variants of all our aspects in their policies \citep{plos-reviewing-guidelines}.
Finally, all reviews in \datasetname are in English only.\looseness=-1

\paragraph{Task Context}
The collected data and trained models only consider the review comment. Preliminarily, we found that the entire review is seldom required to understand a particular review comment. However, some comments may have a higher utility when also considering the parts of the paper they are referring to. In retrospect, we have verified this experimental design choice by running our synthetic data generation pipeline with the full paper provided in the context for 100 random review comments. Results are presented in Appendix~\ref{app:paper_context_exp}. We find that the alignment with human data is better without the full paper context (likely due to the phenomenon that LLMs are easily distracted by too much irrelevant context \citep{shi-et-al-llm-distraction-2023}), except for \verifiability, which could benefit from additional context. We further investigated the label disagreements between the two setups and found that on average, on 89.25\% of samples, the two setups fully agree or deviate in the predicted label by at most $\pm1$.

\paragraph{Segmentation of Reviews into Comments}
While our rule-based approach to segmenting review text into independent comments is mostly effective, it remains imperfect. Some were incorrectly split, resulting in fragments that no longer formed coherent units. For production setups, this segmentation method should be improved. Besides, the four aspects are not always relevant to all extracted comments; for example, trivial comments about the layout or typos, or comments that are formulated as questions, might not always require a high verifiability. In production setups, these comments could be further filtered.

\paragraph{Addressing Different Review Audiences}
Although most review comments are directed toward authors, we recognize that some are intended for meta-reviewers. Consequently, some assessments may not apply to comments aimed at meta-reviewers rather than authors. These often summarize or critique aspects of the paper to assist meta-reviewers in making decisions, rather than guiding authors on how to improve their draft. 

\paragraph{Task subjectivity}
While we do thorough efforts on designing detailed guidelines for each aspect and how they should be labeled, the task remains somewhat subjective. This becomes more prominent in the \helpfulness aspect. It still depends on the annotator's judgment and skills.
Moreover, judging scientific manuscripts requires skill and experience. This makes the human labels heavily dependent on their annotator. 

\section*{Ethical Consideration}
We stress that this work is intended to help reviewers improve their written reviews by providing automatic feedback to them. While we believe that reviewers need to be supported to cope with the rising workload, we emphasize that human reviewers bear the responsibility for their reviews. Reviewers should therefore refrain from acting upon feedback (human or synthetic) blindly. Further, our data annotation did not require collecting any personal data from annotators. Recruitment and all communication were conducted via Prolific.

\section*{Acknowledgments}
We thank the NLP Department at MBZUAI for funding the hiring of human annotators, making this work possible. We thank the reviewers for their helpful suggestions for improving this paper, as well as LU Sheng and Yiwei Wang for their insightful feedback throughout the paper-writing process.

This work has been funded by the LOEWE Distinguished Chair ``Ubiquitous Knowledge Processing'', LOEWE initiative, Hesse, Germany (Grant Number: LOEWE/4a//519/05/00.002(0002)/81), by the European Union (ERC, InterText, 101054961) and by the German Research Foundation (DFG) as part of the PEER project (grant GU 798/28-1). Views and opinions expressed are, however, those of the authors only and do not necessarily reflect those of the European Union or the European Research Council. Neither the European Union nor the granting authority can be held responsible for them.

\bibliography{anthology-1,anthology-2,custom}

\appendix

\section{Detailed Aspects Definitions}
\label{app:aspects_definitions}
\subsection{Actionability}
This aspect measures the actionability of a review comment, i.e., how much it provides useful guidance for the authors. It is important to authors as they seek clear guidance to help them improve their draft.

We evaluate actionability based on two dimensions:
(1) Is the action stated directly, or does the author need to infer it? (Explicit vs. Implicit)
(2) After identifying the action, is it clear how to apply it, or is it vague? (Concrete vs. Vague)

These categories are defined as follows:
\begin{itemize}
    \item \textbf{Explicit:} The comment provides direct or apparent actions or suggestions. Authors can immediately identify the modifications needed for their draft. Clarification questions are considered explicit if they imply a direct action.
    
    \item \textbf{Implicit:} The comment suggests actions indirectly. This can be in the form of questions that need to be addressed or missing parts that need to be added. Actions are not explicitly stated, but authors can infer what needs to be done after reading the comment.
    
    \item \textbf{Concrete:} After identifying the action, the authors clearly understand what needs to be done and how to apply the suggested changes.
    
    \item \textbf{Vague:} Even after identifying the action, the authors still do not know how to carry it out.
\end{itemize}

We note that it is more important for review comments to be \textit{concrete} than to be \textit{explicit}, as authors can often infer implicit actions, but they need the action to be concrete to ensure clarity and effective implementation. This preference is reflected in our scoring schema.

\paragraph{Labels}
Building on these definitions, we define five labels corresponding to five levels of actionability:

    \paragraph{1 - Unactionable:} The comment lacks any meaningful information to help the authors improve the paper. After reading the comment, the authors do not know what they should do. 
    \begin{quote}
    Example:
    The idea of using positional encodings (PEs) for GNNs on molecular graph regression is not new.
    \end{quote}

    \paragraph{2 - Borderline Actionable:} The comment includes an \textit{implicitly} stated action, or the action can be inferred. However, the action itself is \textit{vague} and lacks details on how to apply it.
    \begin{quote}
        Example:
        It is not clear if this trend holds across different model architectures.
    \end{quote}
    
    \paragraph{3 - Somewhat Actionable:} The comment \textit{explicitly} states an action but is \textit{vague} on how to execute it.
    \begin{quote}
        Example:
        You should address the lack of technical novelty in this paper.
    \end{quote}
    \paragraph{4 - Mostly Actionable:} The comment \textit{implicitly} states an action but \textit{concretely} describes how to implement the inferred action.
    \begin{quote}
        Example:
        There are some very relevant baselines like X and Y that other people have been comparing their results to.
    \end{quote}
    \paragraph{5 - Highly Actionable:} The comment contains an \textit{explicit} action and provides \textit{concrete} details on how to implement it. The authors know exactly how to apply the suggested changes.
    \begin{quote}
        Example:
        What will happen if you use the evaluation metric X instead of Y?
    \end{quote}

\subsection{Grounding and Specificity}
This aspect measures how explicitly a review comment is linked to a specific part of the paper. This is crucial for authors, as it helps them identify which section of their paper needs revision. Additionally, it evaluates how clearly the comment specifies the issue with that section.

This aspect consists of two dimensions:\\
(1) \textbf{Grounding}: What part of the paper does the comment address?  
(2) \textbf{Specificity}: What is wrong with this part?

\begin{itemize}
    \item \textbf{Grounding:} Measures how well the authors can identify the part of the paper being addressed by the comment. This can be classified into three levels: no grounding, weak grounding, or full grounding.
    \begin{itemize}
        \item \textbf{Weak grounding:} The authors cannot precisely identify the section being addressed, but they have some hint or an educated guess.
        \item \textbf{Full grounding:} The authors can accurately determine which part of the paper is being referenced. This can be achieved through:
        \begin{itemize}
            \item Explicit mentions of sections, tables, figures, etc.
            \item Discussion of a unique aspect of the paper that allows the authors to infer the reference.
            \item General comments that, while not mentioning specific parts, make it clear which sections are being addressed.
        \end{itemize}
    \end{itemize}
    
    \item \textbf{Specificity:} Measures the level of detail provided in identifying what is wrong or missing in the referenced part of the paper. If the comment refers to external work, specificity also evaluates whether concrete examples are provided.
\end{itemize}

We note that grounding is more important than specificity, as it's more important for the authors to identify the part of the paper that is being addressed by the comment.

\paragraph{Labels}
We define five label categories as follows:\\
    \paragraph{1 - Not Grounded:} The comment is not grounded at all. It does not reference any specific section of the paper and is highly unspecific.
    \begin{quote}
        Example:
        The paper discusses a hot topic in the field now. However, one major drawback to this draft is that the analysis is poor.
    \end{quote}
    \paragraph{2 - Weakly Grounded and Not Specific:} The authors cannot confidently determine which part of the paper the comment addresses, and the comment does not specify what needs to be revised.
    \begin{quote}
        Example:
        For many of the datasets tested, the improvement over other approaches or even the general adversarial approach is marginal. 
    \end{quote}
    \paragraph{3 - Weakly Grounded and Specific:} The authors cannot confidently determine the referenced section, but the comment clearly specifies what needs to be addressed.
    \begin{quote}
        Example:
        Some figures need their captions to be more precise and to define all variables used in the figure.
    \end{quote}
    \paragraph{4 - Fully Grounded and Under-Specific:} The comment explicitly mentions or makes it obvious which part of the paper it addresses, but it does not clearly specify what needs to be revised.
    \begin{quote}
        Example:
        In Figure 7, the results and supplemental video results show that SurfGAN seems out of place.
    \end{quote}
    \paragraph{5 - Fully Grounded and Specific:} The comment explicitly mentions or makes it obvious which part of the paper it addresses, and it clearly specifies what needs to be revised.
    \begin{quote}
        Example:
        The differences in results in Table 2 are very small that make the interpretation of results rather difficult. Furthermore, it is then unclear which proposed methods are really effective.
    \end{quote}

\subsection{Verifiability}
This aspect evaluates whether a review comment contains a claim (i.e., a subjective opinion) and how well it is substantiated. The first step is to determine whether the comment includes any claims. If it does, we assess the extent to which the reviewer justifies or supports the claim through logical reasoning, common knowledge, or references. Justifications can appear before or after the claim, and claims do not always need to be stated explicitly—they can also be inferred.

\paragraph{Opinions \& Claims}
Claims are subjective statements, including:\\
\begin{itemize}
    \item Opinions or stances taken by the reviewer, such as disagreements with an experimental choice.
    \item Suggestions or requests for changes, such as recommendations to add, remove, or discuss a certain topic.
    \item Evaluative judgments about parts of the paper, such as comments on readability, level of detail, or quality.
    \item Deductions or inferred observations that extend beyond simply stating facts or results from the paper.
\end{itemize}

Any statement that requires supporting evidence to help the authors understand the reviewer’s reasoning qualifies as a claim. These can be either direct or indirect.\\
\textbf{Example:} “Important methods like X are not discussed.” This implies that method X should be discussed. The reviewer should explain why discussing this method is necessary.

\paragraph{Verification}
A claim can be substantiated through:\\
\begin{itemize}
    \item \textbf{Logical reasoning}: The reviewer explains the reasoning behind the claim.
    \item \textbf{Common knowledge in the field}: The claim is supported by well-established practices or standards.
    \item \textbf{External references}: The claim is backed by citations or external sources.
\end{itemize}

\paragraph{Normal Statements}
Statements that do not contain claims should be labeled as \texttt{No Claim}. These include:
\begin{itemize}
    \item Statements that simply indicate the presence or absence of something, without suggesting changes.
    \item General comments about the paper that do not express an opinion.
    \item Objective, factual statements that do not require verification.
    \item Clarification requests or general questions.
    \item Logical statements or directly inferable observations.
    \item Positive comments, such as compliments on the paper’s quality, as they offer little actionable feedback for improvement.  \textbf{Example:} “This paper is well written, and the experimentation methods are well designed.”  
    
\end{itemize}

\paragraph{Labels}
Unlike other aspects, verifiability uses a 6-point scale instead of 5, as it includes an additional category for review comments that do not contain claims. The categories are defined as follows:

    \paragraph{1 - Unverifiable:} The comment contains a claim but provides no supporting evidence or justification.
    \begin{quote}
        Example:
        For many of the datasets tested, the improvement over other approaches or even the general adversarial approach is marginal.
    \end{quote}
    \paragraph{2 - Borderline Verifiable:} The comment offers some support for its claim, but the justification is insufficient, vague, or poorly articulated. The authors may struggle to understand the reasoning.
    \begin{quote}
        Example:
        This method shouldn’t achieve good results. If I remember correctly, I have read a paper that tried to do the same thing, but it didn’t work for them.
    \end{quote}
    \paragraph{3 - Somewhat Verifiable:} The comment provides partial support for its claim, but key elements—such as specific examples, detailed explanations, or supporting references—are missing. The justification requires significant effort to follow.
    \begin{quote}
        Example:
        The approximation error is defined as the gap between the objective values, which is somehow ambiguous unless one has seen the values in the table.
    \end{quote}
    \paragraph{4 - Mostly Verifiable:} The claim is sufficiently supported but has minor gaps. The reviewer could enhance the justification by providing more detail or adding references.
    \begin{quote}
        Example:
        The statistical analysis appears incorrect because the p-values reported for the t-tests do not align with standard thresholds for significance.
    \end{quote}
    \paragraph{5 - Fully Verifiable:} The claim is thoroughly supported with explicit, sufficient, and robust evidence. This can be achieved through:
    \begin{itemize}
        \item Clear and precise reasoning or explanation.
        \item Relevant references to external works or data.
        \item Logically sound common-sense arguments.
    \end{itemize}
    
    \begin{quote}
        Example:
      The landscape results in parameter space looks very surprising because it has no assumptions on the generator and discriminator architecture except for enough representation. This looks surprising to me because usually, this kind of global optimization result for neural networks needs strong assumptions on the architecture.
    \end{quote}

    \paragraph{X - No Claim:} The comment does not contain any claim, opinion, or suggestion. It consists solely of factual, descriptive statements that do not require justification.
    \begin{quote}
        Example:
        This algorithm is slow, as it relies on an O(N²) algorithm.
    \end{quote}

\subsection{Helpfulness}
Helpfulness evaluates the overall usefulness of a comment to the authors, serving as an aggregation of the previous three aspects. It is a subjective measure that reflects the value of the review comment. This aspect is rated on a scale from 1 to 5, with the following definitions for each level:

\paragraph{Labels}
    \paragraph{1 - Not Helpful:} The comment fails to identify any meaningful weaknesses or suggest improvements, leaving the authors with no actionable feedback.
    \begin{quote}
        Example:
        In the experiments, the transfer tasks are too artificial.
    \end{quote} 
    
    \paragraph{2 - Barely Helpful:} The comment identifies a weakness or an area for improvement but is vague, unclear, or provides minimal guidance, making it only slightly beneficial to the authors.
     \begin{quote}
        Example:
         Section 5: It is unclear why the superspreader model is more realistic or more challenging than the uniform corruption.
    \end{quote}
    \paragraph{3 - Somewhat Helpful:} The comment points out weaknesses or areas for improvement but lacks depth, specificity, or completeness. While the authors may gain some insights, the feedback does not fully support them in refining their draft.
    \begin{quote}
        Example:
         CRUCIAL: The evaluation is unclear. Were agents evaluated on held-out environments from the same task? Or on the N\_env training environments? Either way seems fine, but it should be specified!
    \end{quote}
    \paragraph{4 - Mostly Helpful:} The comment provides clear and actionable feedback on weaknesses and areas for improvement. However, it could be further expanded or refined to be fully comprehensive and impactful.
    \begin{quote}
        Example:
         It is hard to find the formal definition of the proposed CRS model. It seems to be the equation after line 175, but the authors did not say it explicitly.
    \end{quote}
    \paragraph{5 - Highly Helpful:} The comment thoroughly identifies weaknesses and offers detailed, actionable, and constructive suggestions, enabling the authors to make significant improvements to their draft.
    \begin{quote}
        Example:
          The abstract should act like a compact summary of your draft. The way it is not, it needs extra extra summarization. Don’t include a lot of details about your proposed algorithm there.
    \end{quote} 

\section{Review Segmentation}
\label{app:review_segmentation}

The following shows our segmentation process for the reviews.

\begin{itemize}
    \item 	\textbf{Weakness and Questions Sections Extraction:} We begin by extracting the Weaknesses and Questions parts of the review. This is done according to the different venues. Some venues, like ARR, provide each part of the review (summary, strengths, weaknesses) separately. Other venues require string matching to extract those sections.
    
    \item 	\textbf{Text Cleaning:} The second step is to clean the extracted text. We collapse spaces and newline characters and handle some formatting errors due to \LaTeX{} text parsing.
    
    \item 	\textbf{Delimiter-Based Segmentation:} We then apply a complex string-matching script to match most of the delimiters that indicate the beginning of new review comments. We begin by checking if this is the beginning of the sentence, and then match with bullet-points delimiters (for example: *, $\bullet$, numbers, etc). We also adapt to cases, where we have mentioned keywords like (W, Weakness, et).
    
    Some of the supported delimiters are:
    \begin{itemize}
        \item - [COMMENT]
        \item $\bullet$ [COMMENT]
        \item * [COMMENT] 
        \item (W[NUMBER]) [COMMENT]
    \end{itemize}
    
    \item 	\textbf{Merging Short Comments:} After that, we merge comments that have fewer than five words together. 
    
    \item 	\textbf{Filtering Typos:} We also exclude review comments that only mention typo fixes, as these hold low value for the authors.
    
    \item 	\textbf{Only Bullet Points:} In this step, we only consider comments that start with bullet-point delimiters.
    
    \item 	\textbf{Post-Rebuttal Comments:} In this filter, we remove all comments that mention post-rebuttal discussion, as these usually hold low value for the authors.
    
    \item 	\textbf{Final Length Filter:} Finally, we measure the mean ($\mu$) and standard deviation ($\sigma$) of the length of the review comment and remove all comments with length greater than $\text{max\_length}$ and comments with length less than $\text{min\_length}$. These are defined as:
    \begin{align*}
        \text{min\_length} &= \mu - \sigma \\
        \text{max\_length} &= \mu + \sigma
    \end{align*}
    In our dataset, these statistics are: 
    \textbf{Mean:} 53.13 \quad \textbf{Std:} 47.62 \quad \textbf{Min:} 5.51 \quad \textbf{Max:} 100.76 words.
After this process, we exclude all comments that still have a length of fewer than 10 words.
\end{itemize}

We note that excluding some parts of the review, and not considering them, is not a significant problem in our current setting that focuses on creating a rich dataset from high-quality samples. All the scripts and filters we used are included in our code release.

After applying the segmentation process, we obtain a total of 207k unique review comments from weaknesses and discussion sections of reviews. Table~\ref{tab:review_stats} reports the statistics of our raw data across the five venues, from the NLP and ML domains, and over seven years. This dataset is the base that we sample from for our human-labeled (see \S\ref{ssec:human_data}, and synthetic data (see \S\ref{ssec:synthetic_data}) collection.

\begin{table}[ht!]
    \centering
    \resizebox{\linewidth}{!}{ 
    \begin{tabular}{@{}lrrrr@{}}
        \toprule
        \textbf{Venue} & \textbf{\#R} & \textbf{\#P} & \textbf{|R|} & \textbf{|P|} \\
        \midrule
        ACL 2017         & 140 & 542    & 474  & 38  \\
        ARR 2022         & 404 & 1701   & 320  & 38  \\
        EMNLP 2023       & 3156 & 8616  & 157  & 39  \\
        ICLR 2021        & 220 & 591    & 518  & 42  \\
        ICLR 2022        & 544 & 1447   & 393  & 40  \\
        ICLR 2023        & 880 & 2077   & 266  & 39  \\
        ICLR 2024        & 18343 & 63280 & 261  & 42  \\
        ICLR 2025        & 33032 & 119311 & 275 & 43  \\
        NeurIPS 2016     & 41 & 137     & 316  & 39  \\
        NeurIPS 2017     & 69 & 254     & 315  & 39  \\
        NeurIPS 2018     & 227 & 895    & 439  & 40  \\
        NeurIPS 2019     & 146 & 516    & 482  & 43  \\
        NeurIPS 2020     & 2354 & 6635  & 224  & 43  \\
        NeurIPS 2021     & 171 & 445    & 417  & 43  \\
        NeurIPS 2022     & 357 & 740    & 249  & 38  \\
        \midrule
        \textbf{Total} & \textbf{60084} & \textbf{207187} & \textbf{341}  & \textbf{40}  \\
        \bottomrule
    \end{tabular}
    }
    \caption{Number of reviews (\textbf{\#R}) and extracted, individual review comments (\textbf{\#P}) across venues and years. The |R| and |P| columns report the average number of words, rounded to the nearest integer.}
    \label{tab:review_stats}
\end{table}

\section{Annotation Guidelines}
\label{ssec:annotation_guidelines}
To formally introduce the task and familiarize human annotators with it, we design an annotation guideline in the form of a questionnaire. We used the Tally\footnote{\url{https://tally.so}} platform to create a form that includes both definitions and questions to assess the workers' understanding of the task. We pay annotators 16\$ per hour, where 12\$ went to the annotators, and 4\$ to the annotation platform. We informed all participants of the scope of the project and collected their consent to use the data. 

To carry out the annotation process, we utilize Argilla\footnote{\url{https://argilla.io}}, which we deploy on a Hugging Face Space\footnote{\url{https://huggingface.co/spaces}}.

\section{Data Samples}
Table~\ref{tab:data_samples} presents sample annotations across venues of the \datasetname dataset.
\label{app:data-samples}

\begin{table*}[b]
\begin{adjustbox}{angle=90, width=\textwidth} 
\tiny
\begin{tabularx}{0.99\textheight}{@{}p{1cm}XccccccccXXXX@{}}
\toprule
& & \multicolumn{2}{c}{\textbf{Action.}} & \multicolumn{2}{c}{\textbf{Ground.}} & \multicolumn{2}{c}{\textbf{Verify.}} & \multicolumn{2}{c}{\textbf{Help.}} & \textbf{Actionability Rationale} & \textbf{Grounding Rationale} & \textbf{Verifiability Rationale} & \textbf{Helpfulness Rationale} \\
\cmidrule(lr){3-4} \cmidrule(lr){5-6} \cmidrule(lr){7-8} \cmidrule(lr){9-10}
\textbf{Venue} & \textbf{Review Comment} & \textbf{H} & \textbf{S} & \textbf{H} & \textbf{S} & \textbf{H} & \textbf{S} & \textbf{H} & \textbf{S} & & & & \\
\midrule

ICLR & Adding a method on the top of other methods to improve transferability is good but cannot be considered a significant contribution. & 1 & 1 & 1 & 1 & 1 & 1 & 1 & 1 & The review point comments on the addition of a method to improve transferability, stating that while it is beneficial [..] However, it does not provide any explicit or implicit actions for the authors to take. [..] this comment is unactionable.
 & The comment critiques the significance of adding a method on top of existing methods to improve transferability, suggesting it may not be a substantial contribution. [..] this comment aligns with a score of 1.
 & The review point claims that adding a method on top of other methods to improve transferability is not a significant contribution. However, it does not provide any supporting evidence, [..] the comment is considered unverifiable.
 & The review comment critiques the significance of the contribution by stating that adding a method on top of existing methods to improve transferability is not a significant contribution. [..] The comment is not helpful at all. \\

\midrule

NeurIPS & The proposed method uses AdamW with cosine lr for training, while comparing methods only use adam with fixed lr. Directly comparing with their numbers in paper is unfair. It would be better to reproduce their results using the same setting, since most of the recent methods have their code released. & 5 & 5 & 5 & 3 & 4 & 3 & 5 & 4 & The review comment highlights a potential issue with the fairness of the comparison between the proposed method and other methods due to differences in the training setup, [..]  The explicit nature of the suggestion and the detailed guidance on how to implement it make this comment highly actionable.
 & [..] The suggestion is specific, as it outlines a particular approach to improve the fairness of the comparison. Therefore, the comment is weakly grounded but specific, aligning with the label 3.
 & The review point claims that the comparison between the proposed method and others is unfair due to differences in training configurations,[..]  but it lacks specific examples or references to substantiate the impact of these differences. [..]  the comment is rated as "3: Somewhat Verifiable."
 &  [..] This feedback is clear and actionable, providing the authors with a concrete step to enhance the validity of their comparisons. However, it could be more helpful if it included additional guidance on how to implement this suggestion or the potential impact of these differences on the results. Overall, the comment is mostly helpful, aligning with a score of 4. \\

\midrule

NeurIPS & - l148: known, instead of know - l156: please define \textbackslash gamma\_0\textasciicircum{***} - Figure 1: Please specify the meaning of the colors in the caption as well as the text. & 5 & 5 & 5 & 5 & \textbf{X} & \textbf{X} & 5 & 5 & [..]  It suggests correcting a typo in line 148, defining a term in line 156, and specifying the meaning of colors in Figure 1. Each of these points is specific and actionable, giving the authors clear guidance on how to improve their draft. [..] Therefore, this comment is rated as a 5 on the actionability scale.
 & The comment is fully grounded as it explicitly mentions specific locations in the paper, such as "l148," "l156," and "Figure 1," allowing the authors to accurately identify the parts of the paper being addressed. It is also specific because it details what needs to be corrected or clarified, such as correcting a typo, defining a term, [..] aligning with category 5.
 & The review point consists of factual statements and requests for clarification or corrections. It does not contain any subjective opinions, claims, or suggestions that require justification or evidence. [..] it fits the classification of "X"
 & The review comment provides specific and actionable feedback on several minor issues in the paper.  [..]  These precise corrections and requests for clarification are valuable for improving the clarity and accuracy of the manuscript. [..] aligns with a score of 5. \\

\midrule

EMNLP & Details around the filtering process used to create the Arabic climate change QA dataset are lacking. More information on the translation and filtering methodology is needed to assess the dataset quality. & 3 & 3 & 3 & 3 & 3 & 3 & 3 & 3 & The review comment identifies a specific area where additional information is needed: the filtering process used to create the Arabic climate change QA dataset.  [..] While the comment clearly indicates what information is missing, [..] the action is explicit but somewhat vague in terms of execution, making the comment somewhat actionable.
 & The comment identifies a specific aspect of the paper related to the filtering process used to create the Arabic climate change QA dataset. [..]  the comment does not explicitly mention which section of the paper discusses this process, making it weakly grounded. The authors might infer that it relates to the methodology or dataset description sections, [..] the comment is weakly grounded but specific, aligning with a score of 3.
 & The review point claims that details about the filtering process for creating the Arabic climate change QA dataset are insufficient, [..] the comment does not provide specific examples or references to support why these details are crucial or how their absence impacts the assessment of dataset quality. This lack of detailed justification makes the claim somewhat verifiable, [..] 
 & [..]  the comment highlights a crucial aspect that could affect the assessment of the dataset's quality. However, the feedback could be more helpful if it provided specific suggestions on what details should be included or how the authors might present this information.  [..] the comment is evaluated as somewhat helpful, consistent with a score of 3. \\

\midrule
NeurIPS & As I just mentioned, the paper only analyzed, under which cases will the Algorithm 1 converges to permutations as local minima. However, it will be better if the quality of this kind of local minima could be analyzed (e.g. the approximation ratio of these local minima, under certain assumptions). & 5 & 3 & 5 & 5 & \textbf{X} & 3 & 4 & 4 & The review comment suggests that the paper should analyze the quality of local minima, specifically the approximation ratio of these local minima under certain assumptions. [..] The action is explicit, but the details on execution are somewhat vague, making the comment somewhat actionable. & The comment is fully grounded as it explicitly mentions "Algorithm 1," allowing the authors to accurately identify the part of the paper being addressed [..] aligning with category 5. & The review point suggests that the paper's analysis is limited to cases where Algorithm 1 converges to permutations as local minima. It further recommends analyzing the quality of these local minima, such as their approximation ratio under certain assumptions.  [..] The suggestion is somewhat verifiable, as it provides a logical direction for further exploration but does not offer detailed justification or evidence [..] & [..] This feedback is actionable as it provides a clear direction for the authors to enhance their analysis and strengthen their paper. However, the comment could be more helpful if it offered additional guidance or examples on how to conduct this analysis or what specific assumptions to consider. [..] it aligns with a score of 4. \\

\midrule
ICLR & fig 8 shows images with different focusing distance, but it only shows 1m and 5m, which both exist in the training data. How about focusing distance other than those appeared in training? does it generalize well? & 5 & 2 & 5 & 5 & \textbf{X} & \textbf{X} & 5 & 3 & The review comment raises a question about the generalization of the model to focusing distances not present in the training data. It explicitly points out that Figure 8 only shows [..] While the comment clearly identifies an area for improvement, it does not provide specific guidance on how to address this issue, [..]  the comment is barely actionable. & The comment explicitly mentions "fig 8," providing full grounding as the authors can accurately identify the part of the paper being addressed. It specifies the issue by pointing out that the figure  [..] corresponding to label 5. & The comment primarily consists of observations and questions regarding the content of Figure 8. It does not present any claims, opinions, or suggestions that require verification. The focus is on seeking clarification and additional information [..] aligns with the criteria for normal statements. Therefore, it is labeled as "X" & [..] By asking about other focusing distances, the comment suggests an area for further exploration and testing, which could enhance the robustness and applicability of the model. However, the comment could be more helpful if it provided specific suggestions on how to test or demonstrate this generalization capability. Overall, the feedback is somewhat helpful as it identifies a potential gap in the analysis but lacks detailed guidance on addressing it. \\

\midrule

NeurIPS & The experimental comparisons are not enough. Some methods like MoCo and SimCLR also test the results with wider backbones like ResNet50 (2×) and ResNet50 (4×). It would be interesting to see the results of proposed InvP with these wider backbones. & 5 & 4 & 5 & 3 & 5 & 3 & 5 & 3 & The reviewer suggests that the experimental comparisons are insufficient [..] While the comment implies a specific action, it does not explicitly instruct the authors to conduct these tests. However, the suggestion is concrete, as it specifies the exact backbones to be used for comparison. Therefore, the comment is mostly actionable [..]
& The comment suggests expanding the experimental comparisons by testing the proposed InvP method with wider backbones like ResNet50 (2×) and ResNet50 (4×) [..]  This aligns with a score of 3. & The review point claims that the experimental comparisons are insufficient, suggesting that methods like MoCo and SimCLR also test results with wider backbones such as ResNet50 (2×) and ResNet50 (4×). [..]  the comment is classified as 3: Somewhat Verifiable. & The review comment identifies a gap in the experimental comparisons by noting that other methods, such as MoCo and SimCLR, have tested results with wider backbones like ResNet50 (2×) and ResNet50 (4×). [..]  the comment is rated as somewhat helpful. \\

\bottomrule
\end{tabularx}
\end{adjustbox}

\caption{Data samples from \datasetname dataset with Human (H) majority and Synthetic (S) labels, in addition to the synthetically generated rationales. "X" refers to the "No Claim" category. }
\label{tab:data_samples}
\end{table*}
\clearpage

\section{Data License}

The data sources reported in~\ref{subsec:raw_data} are with open licenses. Specifically, NLPeer is distributed under CC-BY-NC-SA 4.0  license. EMNLP 2023, and ICLR 2024-2025 reviews have CC-BY 4.0 license. Reviewer2 is distributed under an open Creative Commons license. We also release our dataset with CC-BY-NC-SA 4.0, an open license, to allow future use and utilization of our dataset.

\section{Manual Data Statistics}
\label{app:manual_data_statistics}
\begin{table}[h]
    \centering
    \resizebox{\linewidth}{!}{  
    \begin{tabular}{@{}lrrrr@{}}
        \toprule
        \textbf{Venue} & \textbf{\#R} & \textbf{\#P} & \textbf{|R|} & \textbf{|P|} \\
        \midrule
        ACL 2017      & 43   & 61   & 483  & 42  \\
        ARR 2022      & 77   & 89   & 335  & 43  \\
        EMNLP 2023    & 110  & 113  & 175  & 39  \\
        ICLR 2021     & 64   & 82   & 436  & 44  \\
        ICLR 2022     & 78   & 92   & 379  & 42  \\
        ICLR 2023     & 91   & 97   & 260  & 42  \\
        ICLR 2024     & 116  & 116  & 268  & 49  \\
        ICLR 2025     & 120  & 120  & 290  & 42  \\
        NeurIPS 2016  & 39   & 94   & 306  & 41  \\
        NeurIPS 2017  & 49   & 89   & 316  & 44  \\
        NeurIPS 2018  & 76   & 94   & 436  & 42  \\
        NeurIPS 2019  & 65   & 88   & 419  & 48  \\
        NeurIPS 2020  & 117  & 118  & 209  & 45  \\
        NeurIPS 2021  & 65   & 88   & 401  & 54  \\
        NeurIPS 2022  & 75   & 89   & 223  & 41  \\
        \midrule
        \textbf{Total} & \textbf{1185} & \textbf{1430} & \textbf{329}  & \textbf{44}  \\
        \bottomrule
    \end{tabular}}
    \caption{Number of reviews (\textbf{\#R}) and \textit{human annotated} individual review comments (\textbf{\#P}) across venues and years. The |R| and |P| columns report the average number of words, rounded to the nearest integer.}
    \label{tab:human_labels_stats}
\end{table}

Table~\ref{tab:human_labels_stats} presents statistics on the manually labeled dataset.

In Fig.~\ref{fig:human_labels_dist}, we show the distribution of \golddata, \silverdata, and \harddata labels across different aspects.

\begin{figure}
    \centering
    \includegraphics[width=\linewidth]{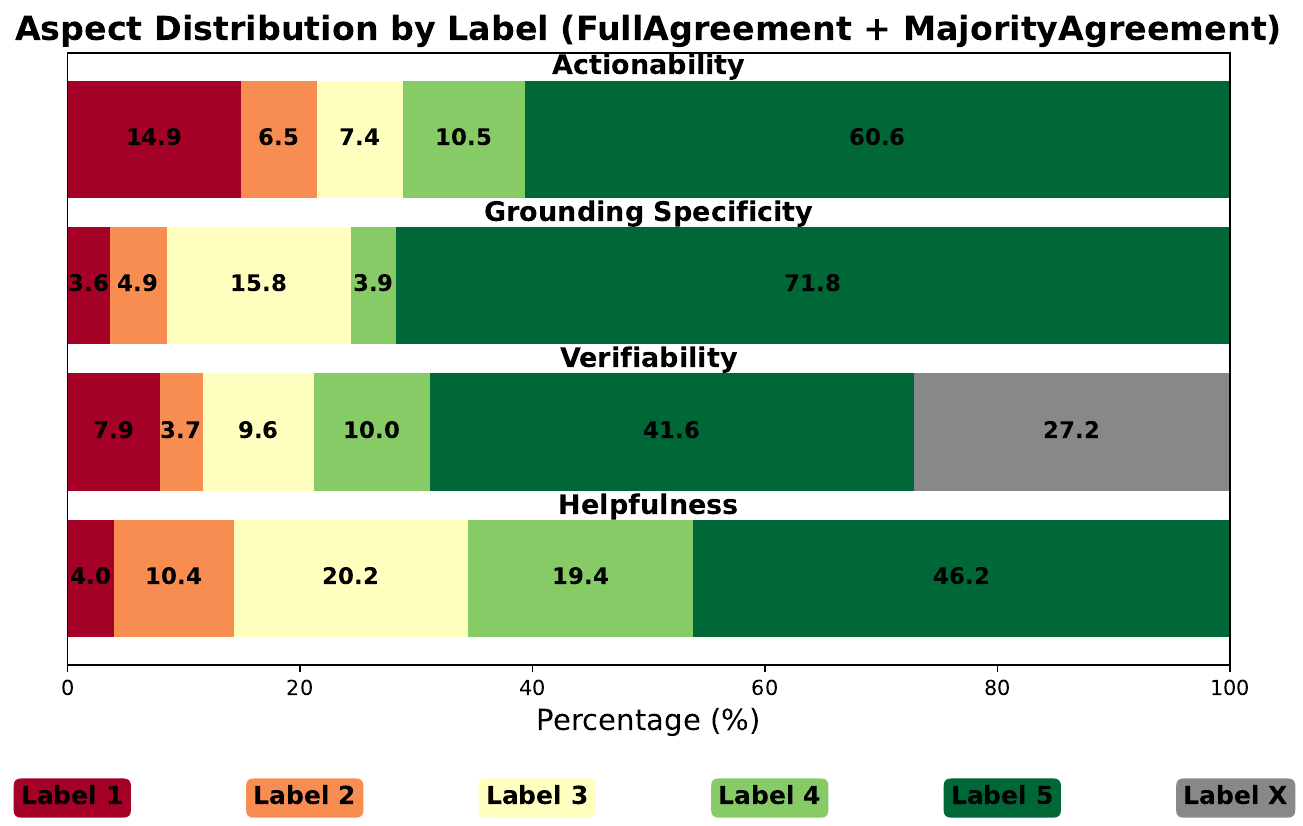}
    \caption{Distribution of the different scores for samples with majority voting (\golddata + \silverdata).}
    \label{fig:gold_silver_dist}
\end{figure}

\begin{figure}
    \centering
    \includegraphics[width=1\linewidth]{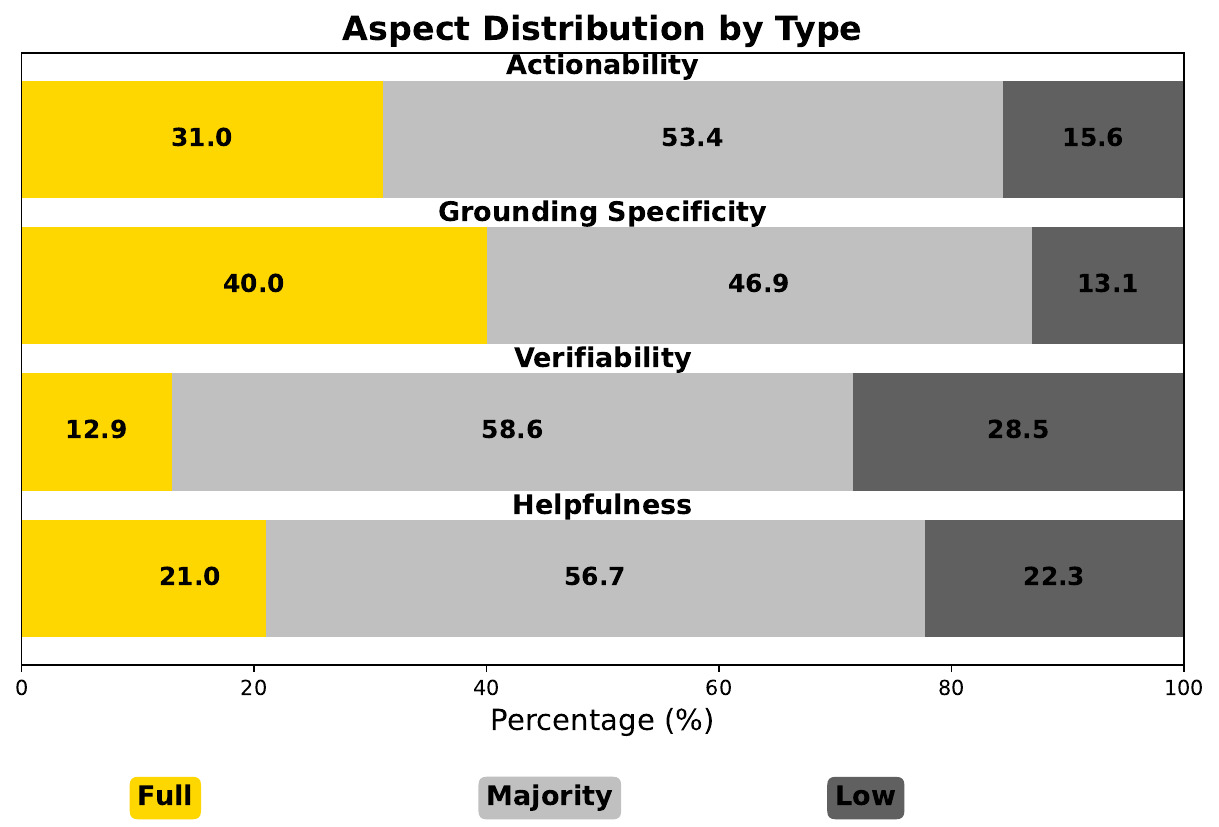}
    \caption{The distribution of label types (\golddata, \silverdata, and \harddata) for the Main four aspects.}
    \label{fig:human_labels_dist}
\end{figure}

To further analyze the distribution of different scores, Fig.~\ref{fig:gold_silver_dist} presents the distribution of \golddata and \silverdata labels. We observe that most labels are concentrated around scores ``5'' and ``1'', indicating that extreme cases are relatively easier for annotators to agree on, while middle-range cases show lower agreement.

\subsection{Detailed Inter Annotator Agreements}
\label{app:detailed_iaa}
In Table~\ref{tab:human_annotators_agreement_pairwise}, we show the detailed agreement details between the three annotators. We note that the low agreement with Annotator Y, on the F1, only leads to underestimating the agreement scores.
\begin{table}[]
\centering

\resizebox{\linewidth}{!}{

\begin{tabular}{@{}llccc@{}} 
\toprule
\textbf{Aspect}                        & \textbf{Annotator Pair} & $\kappa^2$ & $\rho$ & F1  \\ 
\midrule
\multirow{3}{*}{\actionability}         & Z vs X                  & 0.684             & 0.686             & -               \\
                                       & Z vs Y                  & 0.614             & 0.576             & -               \\
                                       & X vs Y                  & 0.543             & 0.531             & -               \\ 
\midrule
\multirow{3}{*}{\groundingshort } & Z vs X                  & 0.402             & 0.409             & -               \\
                                       & Z vs Y                  & 0.465             & 0.432             & -               \\
                                       & X vs Y                  & 0.438             & 0.414             & -               \\ 
\midrule
\multirow{3}{*}{\verifiability}         & Z vs X                  & 0.486             & 0.484             & 0.404           \\
                                       & Z vs Y                  & 0.557             & 0.552             & 0.086           \\
                                       & X vs Y                  & 0.441             & 0.400             & 0.078           \\ 
\midrule
\multirow{3}{*}{\helpfulness}           & Z vs X                  & 0.534             & 0.496             & -               \\
                                       & Z vs Y                  & 0.538             & 0.519             & -               \\
                                       & X vs Y                  & 0.461             & 0.419             & -               \\
\bottomrule
\end{tabular}

}
\caption{Agreement results for the human annotations. The agreement scores are calculated pair-wise between the three annotators. $\kappa^2$ refers to quadratic weighted Kappa. $\rho$ refers to Spearman correlation. F1 refers to the performance of the claim detection.}
\label{tab:human_annotators_agreement_pairwise}
\end{table}

\subsection{Aspect Scores per Venue and Year}
\begin{table}[]
\centering

\resizebox{\linewidth}{!}{
\begin{tabular}{@{}lcccc@{}}
\toprule
\textbf{Venue/Year}   & \textbf{\texttt{A}} & \textbf{\texttt{G\&S}} & \textbf{\texttt{V}} & \textbf{\texttt{H}} \\ 
\midrule
ACL 2017     & 4.31          & 4.73             & 4.10    & 4.18  \\
ARR 2022     & 4.44          & 4.59             & 4.46    & 4.21  \\
EMNLP 2023   & 3.48          & 4.25             & 3.78    & 3.62  \\
ICLR 2021    & 4.07          & 4.47             & 4.08    & 4.15  \\
ICLR 2022    & 3.58          & 4.03             & 3.48    & 3.67  \\
ICLR 2023    & 3.83          & 4.31             & 3.87    & 3.70  \\
ICLR 2024    & 3.82          & 4.43             & 4.20    & 4.02  \\
ICLR 2025    & 3.62          & 4.37             & 4.09    & 3.86  \\
NeurIPS 2016 & 4.36          & 4.46             & 4.16    & 4.13  \\
NeurIPS 2017 & 4.08          & 4.41             & 4.32    & 4.01  \\
NeurIPS 2018 & 4.16          & 4.35             & 3.70    & 3.96  \\
NeurIPS 2019 & 4.19          & 4.59             & 4.54    & 4.15  \\
NeurIPS 2020 & 3.87          & 4.24             & 3.97    & 3.81  \\
NeurIPS 2021 & 4.29          & 4.36             & 4.30    & 4.28  \\
NeurIPS 2022 & 3.45          & 3.84             & 3.33    & 3.45  \\ \bottomrule
\end{tabular}
}
\caption{Average scores per aspect on the \golddata and \silverdata data. \texttt{A} corresponds to \actionability, \texttt{G\&S} to \grounding, \texttt{V} to \verifiability and \texttt{H} to \helpfulness}
\label{tab:aspect_per_year_venue}
\end{table}
Table~\ref{tab:aspect_per_year_venue} shows the average aspect scores per venue, year of our \golddata and \silverdata data. 

\section{Prompts}
\label{app:prompts}
We show the prompts used for generating the synthetic data in Fig.~\ref{fig:synthetic_labels_generation_prompt} and for fine-tuning in Fig.~\ref{fig:finetuning_prompt}.

\begin{figure}[h!]
\begin{tcolorbox}[colback=gray!5!white, colframe=CadetBlue, title=Synthetic Data Labeling Prompt, fonttitle=\bfseries, sharp corners=south]
\scriptsize
This aspect is aimed to maximize the utilization of the review comments for the authors. The primary purpose of the review is to help/guide authors in improving their drafts. Keep this in mind while evaluating the review point. Whenever you encounter a borderline case, think: “Will this review point help authors improve their draft?”. There is no correlation between the aspect score and the length of the review point.

Evaluate the review point based on the aspect description provided next.

[ASPECT]

[ASPECT DESCRIPTION]

Generate a rationale and use it to output the score. 

[INCONTEXT EXAMPLES]

Review Point: [REVIEW POINT]
\end{tcolorbox}
\caption{The prompt used to generate the synthetic labels using GPT-4o}
\label{fig:synthetic_labels_generation_prompt}
\end{figure}

\begin{figure}[h!]
\begin{tcolorbox}[colback=gray!5!white, colframe=CadetBlue, title=Fine-Tuning Prompt, fonttitle=\bfseries, sharp corners=south]
\scriptsize
\#\#\#Task Description:
You are an expert in evaluating peer review comments with respect to different aspects. These aspects are aimed to maximize the utilization of the review comments for the authors. The primary purpose of the review is to help/guide authors in improving their drafts. Keep this in mind while evaluating the review point. Whenever you encounter a borderline case, think: “Will this review point help authors improve their draft?”. There is no correlation between the aspect score and the length of the review point.

Aspect: actionability 

[ACTIONABILITY DEFINITIONS]

Aspect: Grounding \& Specificity

[GROUNDING SPECIFICITY DEFINITIONS]

Aspect: Verifiability 

[VERIFIABILITY DEFINITIONS]

Aspect: Helpfulness 

[HELPFULNESS DEFINITIONS]

\#\#\#Instruction:

Evaluate the review based on the given definitions of the aspect(s) above. Generate a rationale and use it to output the score. Escape the double qoutes inside the rationale.

\#\#\#Review Point: [REVIEW POINT]

\#\#\#Output: \{
  
  "actionability\_rationale": "[ACTIONABILITY RATIONALE]",
  
  "actionability\_label": "[ACTIONABILITY LABEL]",
  
  "grounding\_specificity\_rationale": "[GROUNDING SPECIFICITY RATIONALE]", 
  
  "grounding\_specificity\_label": "[GROUNDING SPECIFICITY LABEL]",
  
  "verifiability\_rationale": "[VERIFIABILITY RATIONALE]", 
  
  "verifiability\_label": "[VERIFIABILITY LABEL]",
  
  "helpfulness\_rationale": "[HELPFULNESS RATIONALE]", 
  
  "helpfulness\_label": "[HELPFULNESS LABEL]"

\}
\end{tcolorbox}
\caption{The prompt used for fine-tuning.}
\label{fig:finetuning_prompt}
\end{figure}

\section{Synthetic Data Statistics and Distribution}
\label{app:synthetic_data_statistics}
In total, we labeled 10,000 review comments across the four aspects. Table~\ref{tab:synthetic_data_stats} provides an overview of the dataset statistics. On average, each review contains about 1.33 labeled comments.

\begin{figure}
    \centering
    \includegraphics[width=1\linewidth]{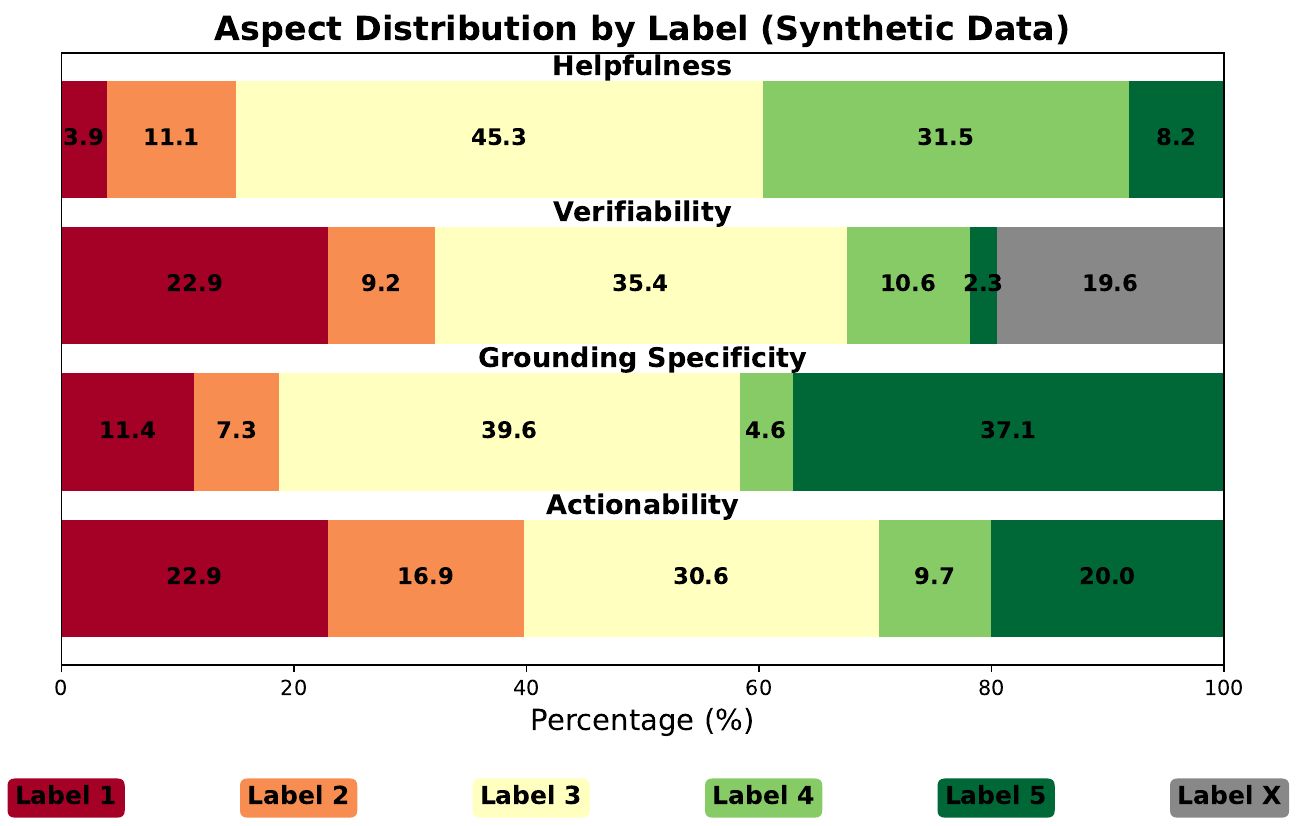}
    \caption{Distribution of the different scores for the synthetic dataset.
    }
    \label{fig:synthetic_data_dist}
\end{figure}

Fig.~\ref{fig:synthetic_data_dist} illustrates the distribution of synthetic labels. The most frequently predicted label is "3," followed by "5" for \actionability and \grounding, and "4" for helpfulness. Compared to human-labeled data, the synthetic label distributions appear more uniform, suggesting a more balanced representation across the different categories, which shows some divergence from the human judgments. We note that we don't have ground truth for the synthetic labels; however, the synthetic labels on the \golddata and \silverdata datasets show high agreement with human labels (see \S\ref{ssec:synthetic_data}) while being generated using the same method used with the synthetic labels here.

\begin{table}[]
    \centering
    \resizebox{\linewidth}{!}{
    \begin{tabular}{@{}lrrrr@{}}
        \toprule
        \textbf{Venue} & \textbf{\#R} & \textbf{\#P} & \textbf{|R|} & \textbf{|P|} \\ 
        \midrule
        ACL 2017     & 104  & 296  & 471 & 39 \\
        ARR 2022     & 317  & 714  & 306 & 41 \\
        EMNLP 2023   & 785  & 897  & 152 & 39 \\
        ICLR 2021    & 139  & 285  & 499 & 46 \\
        ICLR 2022    & 334  & 659  & 350 & 43 \\
        ICLR 2023    & 450  & 688  & 217 & 41 \\
        ICLR 2024    & 1580 & 1652 & 250 & 43 \\
        ICLR 2025    & 2415 & 2519 & 255 & 44 \\
        NeurIPS 2017 & 42   & 66   & 291 & 40 \\
        NeurIPS 2018 & 198  & 561  & 437 & 41 \\
        NeurIPS 2019 & 116  & 257  & 457 & 44 \\
        NeurIPS 2020 & 693  & 825  & 226 & 44 \\
        NeurIPS 2021 & 106  & 188  & 386 & 45 \\
        NeurIPS 2022 & 236  & 393  & 223 & 41 \\
        \midrule
        \textbf{Total} & \textbf{7515} & \textbf{10000}  & \textbf{323} & \textbf{42} \\ 
        \bottomrule
    \end{tabular}
    }
    \caption{Number of reviews (\textbf{\#R}) and \textit{synthetically annotated} individual review points (\textbf{\#P}) across venues and years. The |R| and |P| columns report the average number of words, rounded to the nearest integer.}
    \label{tab:synthetic_data_stats}
\end{table}

\section{Fine-tuning and Evaluation Details}
\label{app:details}

Fine-tuning and Evaluation are done on 1-4 A100 GPUs with 40GB of VRAM each. 
\subsection{Fine-Tuning}

\paragraph{Setup and Hyperparameters}
For fine-tuning, we rely on the HF Alignment Handbook~\cite{tunstall2023zephyrdirectdistillationlm}\footnote{\url{https://github.com/huggingface/alignment-handbook}}. We use this framework as a foundation, modifying it to suit our specific use case. To enable multi-GPU support, we incorporate both the HF Accelerate library\footnote{\url{https://github.com/huggingface/accelerate}} and DeepSpeed-ZeRo~\cite{rajbhandari2020zeromemoryoptimizationstraining}.
All fine-tuning runs were conducted for one epoch using the `bfloat16` datatype~\cite{kalamkar2019studybfloat16deeplearning}. To accelerate training, we employed FlashAttention2~\cite{dao2023flashattention2fasterattentionbetter}. The learning rate was set to $5 \times 10^{-5}$. For experiments involving score-only generation, we used a maximum sequence length of 3072 tokens, while for those involving both scores and rationales, we extended the sequence length to 4096 tokens.
For all fine-tuning experiments, we utilized LoRA with a rank (\texttt{lora\_r}) of 32, \texttt{lora\_alpha} of 16, \texttt{lora\_dropout} of 0.1, and applied it to the following target modules: \texttt{q\_proj}, \texttt{k\_proj}, \texttt{v\_proj}, \texttt{o\_proj}, \texttt{gate\_proj}, \texttt{up\_proj}, and \texttt{down\_proj}.

\paragraph{Completion-Only Loss Calculation}
A key optimization that improved model performance was computing loss only on the completion tokens. By default, loss is calculated over all tokens, including prompts, inputs, and outputs. However, in our case, problem instructions and definitions (see \S\ref{app:prompts}) occupy a significant portion of the prompt. Since these instructions and definitions are given as input and do not require learning, restricting the loss computation to the predictions alone enabled the model to learn more efficiently.

\subsection{Evaluation}
For evaluation, we used the vLLM framework~\citep{vllm}. For the BertScore metric, we used the \textit{bert-base-uncased} model.
To evaluate the fine-tuned models, we used the same prompts as in the fine-tuning process. However, for baseline models, we additionally provided a JSON schema to structure the output.

Since our models were trained to generate scores for all four aspects simultaneously, we followed the same approach in the evaluation process by prompting the models to score all four aspects at once.

All fine-tuned models were evaluated using the instruction prompts (see Appendix~\ref{app:prompts}). However, in the zero-shot evaluation, we used the chat template for all models, as these models were instruction-tuned using this template. The only exception was the Llama3 model, for which we retained the instruction prompt since it is a base model rather than an instruction-tuned model. For GPT-4o and Prometheus, we first applied the Instruction template, then applied the chat template.

All evaluations are done using zero temperature. 
For the evaluations, all models had a sequence length of 1024, except for the DeepReviewer baseline, which had 2048, as the model was generating lengthy explanations.

Due to occasional faulty outputs (e.g., missing scores or failure to parse predictions correctly), there were cases where we could not extract all generated scores. In such instances, we computed evaluation results based only on the successfully parsed scores.

\subsection{Training and Inference Times}
Fine-tuning the Llama-3B model for one epoch takes 90 minutes, while the 8B model took 120 minutes. For inference, to predict all aspect scores for a single review comment, it takes on average  0.67s for Llama 3B, 5.0s for Llama 8B, and 10.1s for Llama 70B.

\section{Aspect Score Evaluation}
\label{app:full_results}
Full results tables, containing all evaluated models using the full set of evaluation metrics, are shown in Tables~\ref{tab:full_actionabiilty_results},~\ref{tab:full_grounding_results},~\ref{tab:full_verifiability_results}, and \ref{tab:full_helpfulness_results}.

\begin{table*}
\centering

\resizebox{\textwidth}{!}{

\begin{tabular}{@{}lcccccccccccccc@{}} 
\toprule
& & & \multicolumn{2}{c}{\textbf{Synthetic}}                           & \multicolumn{9}{c}{\textbf{Human}}                                                                          \\ 
\cmidrule(lr){4-5}
\cmidrule(lr){6-14}
\textbf{\textbf{Model}} & \textbf{\textbf{Mode}} & \multicolumn{1}{c}{\textbf{\textbf{Type}}} & $\kappa^2$ & $\rho$ & $\kappa^2_X$ & $\rho_X$ & $\kappa^2_Y$ & $\rho_Y$ & $\kappa^2_Z$ & $\rho_Z$ & $\kappa^2_{avg}$ & $\rho_{avg}$   \\ 
\midrule
\multicolumn{14}{c}{\textbf{Zero-Shot}}                                                                                                                                        \\ 
\hdashline
GPT-4o                  & S                      & Inst.                  & \textbf{0.696} & \textbf{0.702}        & \textbf{0.441} & \textbf{0.494} & \textbf{0.374} & \textbf{0.496} & \textbf{0.375} & \textbf{0.459} & \textbf{0.397} & \textbf{0.483}  \\
GPT-4o                  & S+R                    & Inst.                  & \underline{0.493}  & \underline{0.549}         & 0.316          & \underline{0.406}  & 0.257          & \underline{0.408}  & 0.263          & \underline{0.391}  & 0.279          & \underline{0.402}   \\
Prometheus-2-7B         & S                      & Inst.                  & 0.127          & 0.188                 & 0.071          & 0.114          & 0.068          & 0.142          & 0.087          & 0.159          & 0.075          & 0.138           \\
Prometheus-2-7B         & S+R                    & Inst.                  & 0.309          & 0.364                 & 0.193          & 0.303          & 0.143          & 0.317          & 0.181          & 0.334          & 0.172          & 0.318           \\
Selene-1-Mini-8B        & S                      & Inst.                  & 0.298          & 0.412                 & 0.209          & 0.310           & 0.174          & 0.302          & 0.177          & 0.281          & 0.187          & 0.298           \\
Selene-1-Mini-8B        & S+R                    & Inst.                  & 0.190          & 0.293                 & 0.124          & 0.238          & 0.092          & 0.203          & 0.083          & 0.180          & 0.100          & 0.207           \\
Flow-Judge-v0.1-3.8B    & S                      & Inst.                  & 0.026          & 0.066                 & 0.033          & 0.047          & 0.003          & 0.020          & 0.005          & 0.002          & 0.014          & 0.023           \\
Flow-Judge-v0.1-3.8B    & S+R                    & Inst.                  & 0.163          & 0.215                 & 0.178          & 0.201          & 0.136          & 0.149          & 0.152          & 0.161          & 0.155          & 0.170            \\
Llama-3.1-8B            & S                      & Inst.                  & 0.057          & 0.094                 & 0.002          & -0.001         & 0.031          & 0.067          & 0.032          & 0.070          & 0.022          & 0.045           \\
Llama-3.1-8B            & S+R                    & Inst.                  & 0.155          & 0.173                 & 0.070          & 0.094          & 0.048          & 0.086          & 0.052          & 0.065          & 0.057          & 0.082           \\
Llama-3.1-IT-8B         & S                      & Chat                   & 0.333          & 0.393                 & \underline{0.337}  & 0.365          & \underline{0.315}  & 0.334          & \underline{0.281}  & 0.289          & \underline{0.311}  & 0.329           \\
Llama-3.1-IT-8B         & S+R                    & Chat                   & 0.103          & 0.201                 & 0.074          & 0.146          & 0.055          & 0.129          & 0.052          & 0.112          & 0.060          & 0.129           \\
DeepSeek-R1-7B          & S                      & Chat                   & 0.301          & 0.304                 & 0.225          & 0.238          & 0.182          & 0.223          & 0.168          & 0.198          & 0.192          & 0.220           \\
DeepSeek-R1-7B          & S+R                    & Chat                   & 0.255          & 0.254                 & 0.190          & 0.205          & 0.148          & 0.193          & 0.130          & 0.152          & 0.156          & 0.183           \\
DeepSeek-R1-70B & S   & Chat & 0.426 & 0.550  & 0.324 & 0.364 & 0.334 & 0.341 & 0.282 & 0.300   & 0.313 & 0.335  \\
DeepSeek-R1-70B & S+R & Chat & 0.401 & 0.503 & 0.308 & 0.343 & 0.329 & 0.359 & 0.281 & 0.312 & 0.306 & 0.338  \\
Gemma-3-IT-27B  & S   & Chat & 0.303 & 0.474 & 0.258 & 0.362 & 0.234 & 0.313 & 0.198 & 0.277 & 0.230  & 0.317  \\
Gemma-3-IT-27B  & S+R & Chat & 0.400   & 0.522 & 0.287 & 0.374 & 0.274 & 0.370  & 0.234 & 0.322 & 0.265 & 0.355  \\
SciTülu-7B              & S                      & Chat                   & 0.090          & 0.173                 & 0.053          & 0.155          & 0.037          & 0.146          & 0.039          & 0.128          & 0.043          & 0.143           \\
SciTülu-7B              & S+R                    & Chat                   & 0.145          & 0.169                 & 0.128          & 0.217          & 0.080          & 0.175          & 0.083          & 0.162          & 0.097          & 0.185           \\
SciLitLLM-7B            & S                      & Chat                   & 0.087          & 0.183                 & 0.055          & 0.129          & 0.050          & 0.157          & 0.051          & 0.126          & 0.052          & 0.137           \\
SciLitLLM-7B            & S+R                    & Chat                   & 0.289          & 0.286                 & 0.170          & 0.184          & 0.130          & 0.165          & 0.170          & 0.191          & 0.157          & 0.180           \\
DeepReviewer-7B         & S                      & Chat                   & 0.236          & 0.253                 & 0.217          & 0.226          & 0.190          & 0.209          & 0.184          & 0.189          & 0.197          & 0.208           \\
DeepReviewer-7B         & S+R                    & Chat                   & 0.193          & 0.209                 & 0.202          & 0.205          & 0.163          & 0.177          & 0.150          & 0.161          & 0.172          & 0.181           \\

\midrule
\multicolumn{13}{c}{\textbf{Fine-Tuning}}                                                                                                                                                                                                                   \\ 
\hdashline

Llama-3.2-IT-3B            & S                      & Inst.                  & 0.684          & 0.689 &                0.509	&0.544&	0.413&	0.523&	0.470 &	0.533&	0.464&	0.533        \\

Llama-3.1-8B            & S                      & Inst.                  & 0.724          & 0.730                 & 0.521          & 0.540          & \underline{0.460}  & 0.549          & \underline{0.514}  & \textbf{0.558}          & \underline{0.498}          & 0.549           \\
Llama-3.1-8B            & S+R                    & Inst.                  & 0.705          & 0.715                 & 0.512          & 0.531          & 0.438          & 0.534          & 0.485          & 0.542          & 0.478          & 0.536           \\
Llama-3.1-IT-8B         & S                      & Inst.                  & 0.709          & 0.712                 & \textbf{0.545} & \textbf{0.557} & \textbf{0.470} & \underline{0.555}  & \textbf{0.515} & 0.556  & \textbf{0.510} & \underline{0.556}   \\
Llama-3.1-IT-8B         & S+R                    & Inst.                  & \underline{0.726}  & \textbf{0.738}        & \underline{0.525}  & \underline{0.548}  & 0.452          & \textbf{0.566} & 0.499          & \underline{0.557} & 0.492  & \textbf{0.557}  \\
DeepSeek-R1-7B          & S                      & Inst.                  & 0.532          & 0.533                 & 0.426          & 0.467          & 0.350          & 0.477          & 0.367          & 0.445          & 0.381          & 0.463           \\
DeepSeek-R1-7B          & S+R                    & Inst.                  & \textbf{0.733} & \underline{0.737}         & 0.497          & 0.539          & 0.405          & 0.528          & 0.458          & 0.543          & 0.453          & 0.537           \\
SciTülu-7b              & S                      & Inst.                  & 0.426          & 0.422                 & 0.396          & 0.419          & 0.283          & 0.361          & 0.330          & 0.377          & 0.336          & 0.386           \\
SciTülu-7b              & S+R                    & Inst.                  & 0.674          & 0.682                 & 0.477          & 0.489          & 0.427          & 0.518          & 0.466          & 0.504          & 0.457          & 0.504           \\
SciLitLLM-7B            & S                      & Inst.                  & 0.653          & 0.650                 & 0.492          & 0.505          & 0.424          & 0.493          & 0.447          & 0.478          & 0.454          & 0.492           \\
SciLitLLM-7B            & S+R                    & Inst.                  & 0.664          & 0.674                 & 0.518          & 0.534          & 0.429          & 0.527          & 0.469          & 0.524          & 0.472          & 0.528           \\
DeepReviewer-7B         & S                      & Inst.                  & 0.406          & 0.417                 & 0.295          & 0.353          & 0.164          & 0.226          & 0.195          & 0.252          & 0.218          & 0.277           \\
DeepReviewer-7B         & S+R                    & Inst.                  & 0.565          & 0.577                 & 0.354          & 0.413          & 0.301          & 0.437          & 0.332          & 0.431          & 0.329          & 0.427           \\

\bottomrule
\end{tabular}
}
\caption{Model performance scores of the \actionability aspect across different evaluation settings for the test split of the Synthetic dataset (Synth.), and the full human annotations (Human). We evaluate two generation settings: to only generate a score (S), and to generate a rationale before predicting the score (S+R). $\kappa^2$ refers to quadratic-weighted Cohen's Kappa, $\rho$ to Spearman correlation. The X, Y, Z  subscripts refer to the agreement scores with each of the three annotators. The avg. subscript is the average of all three individual results.}
\label{tab:full_actionabiilty_results}

\end{table*}
\begin{table*}
\centering

\resizebox{\textwidth}{!}{

\begin{tabular}{@{}lcccccccccccccc@{}} 
\toprule
& & & \multicolumn{2}{c}{\textbf{Synthetic}}                           & \multicolumn{9}{c}{\textbf{Human}}                                                                          \\ 
\cmidrule(lr){4-5}
\cmidrule(lr){6-13}
\textbf{\textbf{Model}} & \textbf{\textbf{Mode}} & \multicolumn{1}{c}{\textbf{\textbf{Type}}} & $\kappa^2$ & $\rho$ & $\kappa^2_X$ & $\rho_X$ & $\kappa^2_Y$ & $\rho_Y$ & $\kappa^2_Z$ & $\rho_Z$ & $\kappa^2_{avg}$ & $\rho_{avg}$ \\ 
\midrule
\multicolumn{13}{c}{\textbf{Zero-Shot}}  \\ 
\hdashline
GPT-4o                  & S                      & Inst.                  & \textbf{0.591} & \textbf{0.673}        & \textbf{0.496} & \textbf{0.473} & \textbf{0.514} & \textbf{0.474} & \textbf{0.540} & \textbf{0.565} & \textbf{0.517}          & \textbf{0.504}  \\
GPT-4o                  & S+R                    & Inst.                  & \underline{0.434}  & \underline{0.504}         & \underline{0.473}  & \underline{0.486}  & \underline{0.506}  & \underline{0.452}  & \underline{0.412}  & \underline{0.428}  & \underline{0.464}          & \underline{0.455}   \\
Prometheus-2-7B         & S                      & Inst.                  & 0.109          & 0.113                 & 0.093          & 0.097          & 0.126          & 0.115          & 0.125          & 0.126          & 0.115          & 0.113           \\
Prometheus-2-7B         & S+R                    & Inst.                  & 0.134          & 0.233                 & 0.102          & 0.264          & 0.060          & 0.197          & 0.091          & 0.200          & 0.084          & 0.220           \\
Selene-1-Mini-8B        & S                      & Inst.                  & 0.215          & 0.250                 & 0.183          & 0.193          & 0.229          & 0.232          & 0.178          & 0.180          & 0.197          & 0.202           \\
Selene-1-Mini-8B        & S+R                    & Inst.                  & 0.388          & 0.423                 & 0.351          & 0.421          & 0.294          & 0.380          & 0.338          & 0.371          & 0.328          & 0.391           \\
Flow-Judge-v0.1-3.8B    & S                      & Inst.                  & 0.111          & 0.140                 & 0.104          & 0.126          & 0.076          & 0.087          & 0.145          & 0.155          & 0.108          & 0.123           \\
Flow-Judge-v0.1-3.8B    & S+R                    & Inst.                  & 0.101          & 0.172                 & 0.154          & 0.176          & 0.161          & 0.154          & 0.129          & 0.174          & 0.148          & 0.168           \\
Llama-3.1-8B            & S                      & Inst.                  & 0.155          & 0.200                 & 0.170          & 0.219          & 0.161          & 0.204          & 0.127          & 0.152          & 0.153          & 0.192           \\
Llama-3.1-8B            & S+R                    & Inst.                  & 0.099          & 0.153                 & 0.053          & 0.093          & 0.049          & 0.116          & 0.073          & 0.128          & 0.058          & 0.112           \\
Llama-3.1-IT-8B         & S                      & Chat                   & 0.196          & 0.298                 & 0.271          & 0.304          & 0.349          & 0.314          & 0.194          & 0.199          & 0.271          & 0.272           \\
Llama-3.1-IT-8B         & S+R                    & Chat                   & 0.221          & 0.274                 & 0.310          & 0.314          & 0.298          & 0.253          & 0.252          & 0.248          & 0.287          & 0.272           \\
DeepSeek-R1-7B          & S                      & Chat                   & 0.361          & 0.356                 & 0.290          & 0.325          & 0.228          & 0.289          & 0.264          & 0.273          & 0.261          & 0.296           \\
DeepSeek-R1-7B          & S+R                    & Chat                   & 0.319          & 0.316                 & 0.234          & 0.280          & 0.189          & 0.253          & 0.249          & 0.268          & 0.224          & 0.267           \\
DeepSeek-R1-70B & S   & Chat & 0.343 & 0.493 & 0.413 & 0.472 & 0.435 & 0.431 & 0.380  & 0.458 & 0.409 & 0.454  \\
DeepSeek-R1-70B & S+R & Chat & 0.398 & 0.506 & 0.381 & 0.398 & 0.446 & 0.420  & 0.413 & 0.452 & 0.413 & 0.423  \\
Gemma-3-IT-27B  & S   & Chat & 0.205 & 0.526 & 0.238 & 0.382 & 0.289 & 0.400   & 0.228 & 0.417 & 0.252 & 0.400    \\
Gemma-3-IT-27B  & S+R & Chat & 0.269 & 0.450  & 0.343 & 0.391 & 0.408 & 0.420  & 0.288 & 0.357 & 0.346 & 0.389  \\
SciTülu-7B              & S                      & Chat                   & 0.073          & 0.193                 & 0.039          & 0.183          & 0.022          & 0.128          & 0.028          & 0.098          & 0.030          & 0.136           \\
SciTülu-7B              & S+R                    & Chat                   & 0.141          & 0.211                 & 0.107          & 0.182          & 0.049          & 0.109          & 0.058          & 0.076          & 0.071          & 0.122           \\
SciLitLLM-7B            & S                      & Chat                   & 0.012          & 0.044                 & 0.018          & 0.081          & 0.036          & 0.140          & 0.018          & 0.074          & 0.024          & 0.098           \\
SciLitLLM-7B            & S+R                    & Chat                   & 0.194          & 0.246                 & 0.083          & 0.140          & 0.083          & 0.160          & 0.112          & 0.162          & 0.093          & 0.154           \\
DeepReviewer-7B         & S                      & Chat                   & 0.380          & 0.372                 & 0.247          & 0.262          & 0.180          & 0.229          & 0.276          & 0.293          & 0.234          & 0.261           \\
DeepReviewer-7B         & S+R                    & Chat                   & 0.323          & 0.320                 & 0.261          & 0.286          & 0.200          & 0.251          & 0.297          & 0.316          & 0.253          & 0.284           \\

\midrule
\multicolumn{13}{c}{\textbf{Fine-Tuning}}                                                                                                                                                                                                                   \\ 
\hdashline

Llama-3.2-IT-3B            & S                      & Inst.                  & 0.657 & 0.647 & 0.477 & 0.467 & 0.409 & 0.445 & 0.551 & 0.545 & 0.479 & 0.486 \\

Llama-3.1-8B            & S                      & Inst.                  & 0.636          & 0.618                 & 0.492          & \underline{0.486}  & 0.418          & 0.459          & 0.510          & 0.498          & 0.473          & 0.481           \\
Llama-3.1-8B            & S+R                    & Inst.                  & 0.674          & 0.676                 & 0.494          & 0.481          & 0.408          & 0.430          & 0.543          & 0.542          & 0.482          & 0.484           \\
Llama-3.1-IT-8B         & S                      & Inst.                  & \underline{0.692}  & \underline{0.694}         & \underline{0.499}  & 0.475          & \textbf{0.454}  & \textbf{0.482}  & \textbf{0.625}  & \textbf{0.616}  & \textbf{0.526}  & \textbf{0.524}   \\
Llama-3.1-IT-8B         & S+R                    & Inst.                  & \textbf{0.699} & \textbf{0.700}        & \textbf{0.534} & \textbf{0.503} & \underline{0.449} & \underline{0.462} & 0.568 & \underline{0.565} & \underline{0.517} & \underline{0.510}  \\
DeepSeek-R1-7B          & S                      & Inst.                  & 0.421          & 0.403                 & 0.464          & 0.501          & 0.327          & 0.382          & 0.352          & 0.342          & 0.381          & 0.408           \\
DeepSeek-R1-7B          & S+R                    & Inst.                  & 0.691          & 0.687                 & 0.473          & 0.447          & 0.450          & 0.479          & \underline{0.569}          & 0.568          & 0.497          & 0.498           \\
SciTülu-7B              & S                      & Inst.                  & 0.264          & 0.265                 & 0.305          & 0.383          & 0.206          & 0.292          & 0.219          & 0.237          & 0.243          & 0.304           \\
SciTülu-7B              & S+R                    & Inst.                  & 0.578          & 0.609                 & 0.488          & 0.448          & 0.441          & 0.427          & 0.512          & 0.512          & 0.480          & 0.462           \\
SciLitLLM-7B            & S                      & Inst.                  & 0.548          & 0.532                 & 0.454          & 0.468          & 0.375          & 0.426          & 0.470          & 0.465          & 0.433          & 0.453           \\
SciLitLLM-7B            & S+R                    & Inst.                  & 0.610          & 0.610                 & 0.445          & 0.430          & 0.422          & 0.452          & 0.541          & 0.543          & 0.469          & 0.475           \\
DeepReviewer-7B         & S                      & Inst.                  & 0.300          & 0.315                 & 0.269          & 0.375          & 0.167          & 0.279          & 0.205          & 0.246          & 0.214          & 0.300           \\
DeepReviewer-7B         & S+R                    & Inst.                  & 0.497          & 0.487                 & 0.328          & 0.329          & 0.281          & 0.306          & 0.383          & 0.393          & 0.331          & 0.343           \\
\bottomrule
\end{tabular}

}

\caption{Model performance scores of the \grounding aspect across different evaluation settings for the test split of the Synthetic dataset (Synth.), and the full human annotations (Human). We evaluate two generation settings: to only generate a score (S), and to generate a rationale before predicting the score (S+R). $\kappa^2$ refers to quadratic-weighted Cohen's Kappa, $\rho$ to Spearman correlation. The X, Y, Z  subscripts refer to the agreement scores with each of the three annotators. The avg. subscript is the average of all three individual results.}
\label{tab:full_grounding_results}

\end{table*}

\begin{table*}
\centering

\resizebox{\textwidth}{!}{

\begin{tabular}{@{}lccccccccccccccccccc@{}} 
\toprule
& & & \multicolumn{3}{c}{\textbf{Synthetic}}                           & \multicolumn{13}{c}{\textbf{Human}} \\ 
\cmidrule(lr){4-6}
\cmidrule(lr){7-19}
\textbf{\textbf{Model}} & \textbf{\textbf{Mode}} & \textbf{\textbf{Type}} & $\kappa^2$ & $\rho$ & F1 & $\kappa^2_X$ & $\rho_X$  & $F1_X$ & $\kappa^2_Y$ & $\rho_Y$ & $F1_Y$& $\kappa^2_Z$ & $\rho_Z$ & $F1_Z$& $\kappa^2_{avg}$ & $\rho_{avg}$ &$F1_{avg}$  \\ 
\midrule
\multicolumn{19}{c}{\textbf{Zero-Shot}}                                                                                                                                                                                                                                                                                                                                                     \\ 
\hdashline

GPT-4o                  & S                      & Inst.                  & \underline{0.433}  & \textbf{0.572} & \textbf{0.598} & 0.112          & \underline{0.362} & \textbf{0.101} & 0.171          & \textbf{0.421} & \textbf{0.419} & 0.166          & \underline{0.433}  & \textbf{0.421} & 0.150          & \textbf{0.405} & \textbf{0.314}  \\
GPT-4o                  & S+R                    & Inst.                  & \textbf{0.479} & \underline{0.560}  & \underline{0.304}  & 0.120          & 0.314          & \underline{0.088}  & 0.195          & \underline{0.407}  & \underline{0.173}  & 0.236          & \textbf{0.474} & 0.136  & 0.184          & \underline{0.398}  & 0.132           \\
Prometheus-2-7B         & S                      & Inst.                  & 0.147          & 0.184          & 0.008          & 0.118          & 0.137          & 0.040          & \underline{0.241} & 0.279          & 0.022          & 0.163          & 0.203          & 0.019          & 0.174          & 0.206          & 0.027           \\
Prometheus-2-7B         & S+R                    & Inst.                  & 0.239          & 0.257          & 0.134          & 0.126          & 0.283          & 0.054          & 0.134          & 0.252          & 0.163          & 0.128          & 0.233          & 0.164          & 0.129          & 0.256          & 0.127           \\
Selene-1-Mini-8B        & S                      & Inst.                  & 0.154          & 0.239          & 0.077          & \underline{0.211}          & 0.173          & 0.009          & \textbf{0.353}          & 0.335          & 0.029          & 0.279          & 0.264          & 0.014          & 0.281          & 0.257          & 0.017           \\
Selene-1-Mini-8B        & S+R                    & Inst.                  & 0.444          & 0.478          & 0.093          & 0.158          & \textbf{0.368}          & 0.035          & 0.194          & 0.356          & 0.058          & 0.228          & 0.428          & 0.030           & 0.193          & 0.384          & 0.041           \\
Flow-Judge-v0.1-3.8B    & S                      & Inst.                  & 0.064          & 0.100          & 0.005          & 0.115          & 0.124          & 0.010          & 0.156          & 0.171          & 0              & 0.146          & 0.164          & 0.002          & 0.139          & 0.153          & 0.004           \\
Flow-Judge-v0.1-3.8B    & S+R                    & Inst.                  & 0.026          & 0.058          & 0              & 0.179          & 0.213          & 0              & 0.245          & 0.288          & 0              & 0.247          & 0.297          & 0              & 0.224          & 0.266          & 0               \\
Llama-3.1-8B            & S                      & Inst.                  & 0.095          & 0.133          & 0              & 0.134          & 0.141          & 0              & 0.157          & 0.162          & 0              & 0.153          & 0.173          & 0              & 0.148          & 0.159          & 0               \\
Llama-3.1-8B            & S+R                    & Inst.                  & 0.042          & 0.032          & 0.048          & 0.080          & 0.137          & 0.007          & 0.075          & 0.113          & 0.036          & 0.100          & 0.145          & 0.028          & 0.085          & 0.132          & 0.024           \\
Llama-3.1-IT-8B         & S                      & Chat                   & 0.046          & 0.171          & 0.159          & 0.169          & 0.204          & 0.032          & 0.174          & 0.238          & 0.059          & 0.187          & 0.274          & 0.033          & 0.177          & 0.239          & 0.041           \\
Llama-3.1-IT-8B         & S+R                    & Chat                   & 0.426          & 0.438          & 0.005          & 0.192          & 0.361  & 0.011          & 0.229          & 0.341          & 0.006          & \underline{0.290}  & 0.431          & 0.004          & \underline{0.237}  & 0.378          & 0.007           \\
DeepSeek-R1-7B          & S                      & Chat                   & 0.337          & 0.367          & 0.114          & \textbf{0.220}          & 0.348          & 0.024          & 0.221          & 0.285          & 0.059          & \textbf{0.308} & 0.389          & 0.036          & \textbf{0.250} & 0.341          & 0.040           \\
DeepSeek-R1-7B          & S+R                    & Chat                   & 0.284          & 0.303          & 0.110          & 0.162          & 0.266          & 0.042          & 0.212          & 0.300          & 0.076          & 0.268          & 0.354          & 0.043          & 0.214          & 0.307          & 0.054           \\
DeepSeek-R1-70B & S   & Chat & 0.548 & 0.587 & 0.472 & 0.383 & 0.476 & 0.082 & 0.479 & 0.538 & 0.333 & 0.514 & 0.549 & 0.248 & 0.459 & 0.521 & 0.221  \\
DeepSeek-R1-70B & S+R & Chat & 0.548 & 0.570  & 0.455 & 0.333 & 0.441 & 0.091 & 0.450  & 0.535 & 0.389 & 0.493 & 0.574 & 0.331 & 0.425 & 0.517 & 0.270   \\
Gemma-3-IT-27B  & S   & Chat & 0.339 & 0.453 & 0.417 & 0.257 & 0.276 & 0.114 & 0.471 & 0.483 & 0.260  & 0.414 & 0.448 & 0.243 & 0.381 & 0.402 & 0.206  \\
Gemma-3-IT-27B  & S+R & Chat & 0.390  & 0.499 & 0.193 & 0.352 & 0.353 & 0.067 & 0.505 & 0.507 & 0.161 & 0.540  & 0.571 & 0.136 & 0.466 & 0.477 & 0.121  \\
SciTülu-7B              & S                      & Chat                   & 0.109          & 0.189          & 0              & 0.072          & 0.161          & 0              & 0.109          & 0.177          & 0              & 0.122          & 0.254          & 0              & 0.101          & 0.197          & 0               \\
SciTülu-7B              & S+R                    & Chat                   & 0.185          & 0.207          & 0.030          & 0.134          & 0.222          & 0.010          & 0.103          & 0.151          & 0.034          & 0.135          & 0.181          & 0.030          & 0.124          & 0.185          & 0.025           \\
SciLitLLM-7B            & S                      & Chat                   & 0.069          & 0.148          & 0              & 0.032          & 0.124          & 0.010          & 0.059          & 0.131          & 0              & 0.041          & 0.118          & 0.002          & 0.044          & 0.124          & 0.004           \\
SciLitLLM-7B            & S+R                    & Chat                   & 0.036          & 0.050          & 0.199          & 0.013          & 0.037          & 0.085          & 0.085          & 0.124          & 0.171          & 0.017          & 0.046          & \underline{0.195}  & 0.038          & 0.069          & \underline{0.150}           \\
DeepReviewer-7B         & S                      & Chat                   & 0.253          & 0.262          & 0.230          & 0.157  & 0.213          & 0.070          & 0.235  & 0.292          & 0.167          & 0.214          & 0.278          & 0.162          & 0.202          & 0.261          & \underline{0.133}           \\
DeepReviewer-7B         & S+R                    & Chat                   & 0.277          & 0.279          & 0.248          & 0.177 & 0.265          & 0.078          & 0.223          & 0.277          & 0.167          & 0.223          & 0.302          & 0.140          & 0.208          & 0.281          & 0.128           \\ 

\midrule
\multicolumn{18}{c}{\textbf{Fine-Tuning}}                                                                                                                                                                                                                                                                                                 \\ 
\hdashline

Llama-3.2-IT-3B            & S                      & Inst.                  & 0.648 & 0.649 & \underline{0.611} & \textbf{0.308} & \underline{0.509} & 0.068 & \underline{0.318} & 0.436 & \textbf{0.395} & \textbf{0.443} & \textbf{0.600} & 0.326 & \textbf{0.356} & 0.515 & 0.263 \\

Llama-3.1-8B            & S                      & Inst.                  & 0.640          & 0.634          & 0.599  & 0.286          & \textbf{0.510} & 0.078          & 0.304  & \underline{0.442}  & \underline{0.393} & 0.410  & \underline{0.598} & 0.324          & 0.333  & \underline{0.517}  & 0.265           \\
Llama-3.1-8B            & S+R                    & Inst.                  & \textbf{0.669} & \textbf{0.668} & 0.466          & 0.281          & 0.478          & \underline{0.096}  & 0.300          & 0.430          & 0.350          & 0.406          & 0.572          & \underline{0.328}  & 0.329          & 0.493          & 0.258           \\
Llama-3.1-IT-8B         & S                      & Inst.                  & 0.642          & 0.640          & \textbf{0.644} & \underline{0.299} & 0.500          & 0.084          & \textbf{0.330} & \textbf{0.462} & 0.388  & \underline{0.422} & 0.596  & 0.325          & \underline{0.350} & \textbf{0.519} & \underline{0.266}   \\
Llama-3.1-IT-8B         & S+R                    & Inst.                  & \underline{0.659}  & \underline{0.660}  & 0.439          & 0.264          & 0.465          & 0.094          & 0.284          & 0.405          & 0.310          & 0.408          & 0.582          & 0.274          & 0.319          & 0.484          & 0.226           \\
DeepSeek-R1-7B          & S                      & Inst.                  & 0.581          & 0.586          & 0.475          & 0.271          & 0.485          & 0.088          & 0.254          & 0.371          & 0.344          & 0.375          & 0.563          & 0.296          & 0.300          & 0.473          & 0.243           \\
DeepSeek-R1-7B          & S+R                    & Inst.                  & 0.613          & 0.624          & 0.512          & 0.242          & 0.453          & 0.090          & 0.221          & 0.330          & 0.371          & 0.370          & 0.549          & 0.301          & 0.278          & 0.444          & 0.254           \\
SciTülu-7B              & S                      & Inst.                  & 0.456          & 0.465          & 0.177          & 0.225          & 0.438          & 0.040          & 0.186          & 0.276          & 0.154          & 0.307          & 0.501          & 0.125          & 0.239          & 0.405          & 0.106           \\
SciTülu-7B              & S+R                    & Inst.                  & 0.570          & 0.600          & 0.402          & 0.240          & 0.502          & 0.083          & 0.207          & 0.337          & 0.272          & 0.351          & 0.574          & 0.217          & 0.266          & 0.471          & 0.191           \\
SciLitLLM-7B            & S                      & Inst.                  & 0.581          & 0.583          & 0.508          & 0.288  & 0.506  & \textbf{0.112} & 0.285          & 0.404          & 0.375          & 0.375          & 0.568          & \textbf{0.334} & 0.316          & 0.493          & \textbf{0.274}  \\
SciLitLLM-7B            & S+R                    & Inst.                  & 0.546          & 0.563          & 0.353          & 0.233          & 0.474          & 0.090          & 0.232          & 0.368          & 0.275          & 0.361          & 0.581          & 0.224          & 0.275          & 0.474          & 0.196           \\
DeepReviewer-7B         & S                      & Inst.                  & 0.474          & 0.478          & 0.225          & 0.240          & 0.470          & 0.057          & 0.210          & 0.320          & 0.208          & 0.332          & 0.534          & 0.188          & 0.261          & 0.441          & 0.151           \\
DeepReviewer-7B         & S+R                    & Inst.                  & 0.430          & 0.464          & 0.306          & 0.162          & 0.362          & 0.083          & 0.195          & 0.348          & 0.184          & 0.258          & 0.452          & 0.136          & 0.205          & 0.387          & 0.134           \\

\bottomrule
\end{tabular}
}

\caption{Model performance scores of the \verifiability aspect across different evaluation settings for the test split of the Synthetic dataset (Synth.), and the full human annotations (Human). We evaluate two generation settings: to only generate a score (S), and to generate a rationale before predicting the score (S+R). $\kappa^2$ refers to quadratic-weighted Cohen's Kappa, $\rho$ to Spearman correlation. The X, Y, Z  subscripts refer to the agreement scores with each of the three annotators. The avg. subscript is the average of all three individual results.}
\label{tab:full_verifiability_results}
\end{table*}
\begin{table*}
\centering

\resizebox{\textwidth}{!}{

\begin{tabular}{@{}lcccccccccccccc@{}} 
\toprule
& & & \multicolumn{2}{c}{\textbf{Synthetic}}                           & \multicolumn{9}{c}{\textbf{Human}}                                                                          \\ 
\cmidrule(lr){4-5}
\cmidrule(lr){6-13}
\textbf{\textbf{Model}} & \textbf{\textbf{Mode}} & \multicolumn{1}{c}{\textbf{\textbf{Type}}} & $\kappa^2$ & $\rho$ & $\kappa^2_X$ & $\rho_X$ & $\kappa^2_Y$ & $\rho_Y$ & $\kappa^2_Z$ & $\rho_Z$ & $\kappa^2_{avg}$ & $\rho_{avg}$   \\ 
\midrule
\multicolumn{13}{c}{\textbf{Zero-Shot}}                                                         \\ 
\hdashline
GPT-4o                  & S                      & Inst.                  & \textbf{0.582} & \textbf{0.655}        & \textbf{0.404} & \textbf{0.498} & \textbf{0.346} & \textbf{0.485} & \textbf{0.437} & \textbf{0.551} & \textbf{0.396} & \textbf{0.511}  \\
GPT-4o                  & S+R                    & Inst.                  & \underline{0.530}  & \underline{0.577}         & 0.312          & \underline{0.434}  & 0.254          & \underline{0.385}  & \underline{0.308}  & \underline{0.433}  & \underline{0.291}  & \underline{0.417}   \\
Prometheus-2-7B         & S                      & Inst.                  & 0.318          & 0.370                 & 0.204          & 0.249          & 0.179          & 0.209          & 0.215          & 0.253          & 0.199          & 0.237           \\
Prometheus-2-7B         & S+R                    & Inst.                  & 0.204          & 0.357                 & 0.155          & 0.282          & 0.117          & 0.239          & 0.171          & 0.313          & 0.148          & 0.278           \\
Selene-1-Mini-8B        & S                      & Inst.                  & 0.372          & 0.509                 & 0.299          & 0.364          & 0.292          & 0.317          & 0.283          & 0.345          & 0.291          & 0.342           \\
Selene-1-Mini-8B        & S+R                    & Inst.                  & 0.438          & 0.485                 & 0.245          & 0.346          & 0.171          & 0.263          & 0.197          & 0.292          & 0.204          & 0.300           \\
Flow-Judge-v0.1-3.8B    & S                      & Inst.                  & 0.055          & 0.106                 & 0.063          & 0.073          & 0.020          & 0.029          & 0.035          & 0.045          & 0.039          & 0.049           \\
Flow-Judge-v0.1-3.8B    & S+R                    & Inst.                  & 0.143          & 0.246                 & 0.222          & 0.273          & 0.181          & 0.184          & 0.192          & 0.244          & 0.198          & 0.234           \\
Llama-3.1-8B            & S                      & Inst.                  & 0.155          & 0.171                 & 0.157          & 0.175          & 0.137          & 0.146          & 0.132          & 0.159          & 0.142          & 0.160           \\
Llama-3.1-8B            & S+R                    & Inst.                  & 0.096          & 0.178                 & 0.080          & 0.130          & 0.050          & 0.096          & 0.069          & 0.122          & 0.066          & 0.116           \\
Llama-3.1-IT-8B         & S                      & Chat                   & 0.220          & 0.465                 & 0.241          & 0.374          & 0.273          & 0.317          & 0.227          & 0.358          & 0.247          & 0.350           \\
Llama-3.1-IT-8B         & S+R                    & Chat                   & 0.426          & 0.424                 & \underline{0.337}  & 0.359          & 0.216          & 0.230          & 0.258          & 0.275          & 0.270          & 0.288           \\
DeepSeek-R1-7B          & S                      & Chat                   & 0.387          & 0.419                 & 0.329          & 0.348          & 0.213          & 0.227          & 0.266          & 0.280          & 0.269          & 0.285           \\
DeepSeek-R1-7B          & S+R                    & Chat                   & 0.323          & 0.352                 & 0.281          & 0.303          & 0.214          & 0.258          & 0.253          & 0.283          & 0.249          & 0.281           \\
DeepSeek-R1-70B & S   & Chat & 0.413 & 0.574 & 0.423 & 0.467 & 0.443 & 0.429 & 0.407 & 0.457 & 0.424 & 0.451  \\
DeepSeek-R1-70B & S+R & Chat & 0.444 & 0.553 & 0.412 & 0.427 & 0.447 & 0.439 & 0.437 & 0.470  & 0.432 & 0.445  \\
Gemma-3-IT-27B  & S   & Chat & 0.292 & 0.473 & 0.248 & 0.335 & 0.298 & 0.366 & 0.233 & 0.318 & 0.260  & 0.340   \\
Gemma-3-IT-27B  & S+R & Chat & 0.439 & 0.585 & 0.427 & 0.457 & 0.447 & 0.429 & 0.384 & 0.413 & 0.419 & 0.433  \\
SciTülu-7B              & S                      & Chat                   & 0.085          & 0.101                 & 0.030          & 0.043          & 0.055          & 0.104          & 0.059          & 0.102          & 0.048          & 0.083           \\
SciTülu-7B              & S+R                    & Chat                   & 0.147          & 0.250                 & 0.144          & 0.261          & 0.098          & 0.186          & 0.128          & 0.243          & 0.123          & 0.230           \\
SciLitLLM-7B            & S                      & Chat                   & 0.321          & 0.376                 & 0.156          & 0.180          & 0.117          & 0.123          & 0.143          & 0.172          & 0.139          & 0.158           \\
SciLitLLM-7B            & S+R                    & Chat                   & 0.206          & 0.274                 & 0.131          & 0.172          & 0.100          & 0.132          & 0.137          & 0.180          & 0.123          & 0.161           \\
DeepReviewer-7B         & S                      & Chat                   & 0.362          & 0.374                 & 0.298          & 0.291          & 0.222          & 0.231          & 0.281          & 0.285          & 0.267          & 0.269           \\
DeepReviewer-7B         & S+R                    & Chat                   & 0.342          & 0.362                 & 0.303          & 0.292          & \underline{0.274}  & 0.247          & 0.283          & 0.278          & 0.287          & 0.272           \\

\midrule
\multicolumn{13}{c}{\textbf{Fine-Tuning}}                                                                                                                                                                                                                   \\ 
\hdashline

Llama-3.2-IT-3B            & S                      & Inst.                  & 0.660 & 0.661 & 0.498 & \textbf{0.567} & 0.382 & 0.422 & 0.479 & 0.541 & 0.453 & 0.510
 \\

Llama-3.1-8B            & S                      & Inst.                  & \textbf{0.712} & \textbf{0.704}        & \underline{0.501}  & 0.547  & \underline{0.411}  & \textbf{0.446} & 0.495          & \underline{0.545}  & 0.469          & \underline{0.513}   \\
Llama-3.1-8B            & S+R                    & Inst.                  & \underline{0.683}  & \underline{0.676}         & 0.495          & 0.507          & \textbf{0.426} & 0.431          & \underline{0.500}  & 0.522          & \textbf{0.474} & 0.487           \\
Llama-3.1-IT-8B         & S                      & Inst.                  & 0.681          & \underline{0.676}         & \textbf{0.511} & \underline{0.558} & 0.409          & \textbf{0.446} & \textbf{0.501} & \textbf{0.551} & \textbf{0.474} & \textbf{0.518}  \\
Llama-3.1-IT-8B         & S+R                    & Inst.                  & 0.663          & 0.669                 & 0.488          & 0.503          & 0.397          & 0.392          & 0.481          & 0.512          & 0.455          & 0.469  \\
DeepSeek-R1-7B          & S                      & Inst.                  & 0.539          & 0.544                 & 0.438          & 0.509          & 0.329          & 0.379          & 0.375          & 0.434          & 0.381          & 0.441           \\
DeepSeek-R1-7B          & S+R                    & Inst.                  & 0.652          & 0.645                 & 0.425          & 0.461          & 0.345          & 0.360          & 0.409          & 0.450          & 0.393          & 0.424           \\
SciTülu-7B              & S                      & Inst.                  & 0.486          & 0.533                 & 0.394          & 0.456          & 0.256          & 0.304          & 0.348          & 0.411          & 0.333          & 0.390           \\
SciTülu-7B              & S+R                    & Inst.                  & 0.635          & 0.634                 & 0.465          & 0.490          & 0.363          & 0.392          & 0.426          & 0.472          & 0.418          & 0.451           \\
SciLitLLM-7B            & S                      & Inst.                  & 0.626          & 0.639                 & 0.458          & 0.515          & 0.361          & 0.404          & 0.436          & 0.492          & 0.418          & 0.470           \\
SciLitLLM-7B            & S+R                    & Inst.                  & 0.631          & 0.622                 & 0.449          & 0.492          & 0.357          & 0.393          & 0.416          & 0.464          & 0.407          & 0.450           \\
DeepReviewer-7B         & S                      & Inst.                  & 0.469          & 0.486                 & 0.383          & 0.413          & 0.228          & 0.237          & 0.296          & 0.317          & 0.302          & 0.322           \\
DeepReviewer-7B         & S+R                    & Inst.                  & 0.490          & 0.518                 & 0.306          & 0.371          & 0.218          & 0.278          & 0.261          & 0.330          & 0.262          & 0.326           \\

\bottomrule
\end{tabular}

}

\caption{Model performance scores of the \helpfulness aspect across different evaluation settings for the test split of the Synthetic dataset (Synth.), and the full human annotations (Human). We evaluate two generation settings: to only generate a score (S), and to generate a rationale before predicting the score (S+R). $\kappa^2$ refers to quadratic-weighted Cohen's Kappa, $\rho$ to Spearman correlation. The X, Y, Z  subscripts refer to the agreement scores with each of the three annotators. The avg. subscript is the average of all three individual results.}
\label{tab:full_helpfulness_results}

\end{table*}

\section{Rationale Samples}
\begin{table*}
\centering
\resizebox{\textwidth}{!}{
\begin{tabular}{@{}p{7cm} p{7cm} c c c@{}} 
\toprule
\textbf{Review Comment} & \textbf{Rationale} & \textbf{Aspect} & \textbf{Relevance} & \textbf{Correctness} \\ 
\midrule
The authors should add some details on how to solve the optimization in the main paper. It's an important piece of information currently lacking in the paper. & The review comment explicitly instructs the authors to add details on how to solve the optimization in the main paper, emphasizing that this is an important piece of information currently missing. This direct request provides a clear and specific action for the authors to take, leaving no ambiguity about what needs to be done to improve the draft. The comment is explicit and concrete, making it highly actionable. & \actionability & 5 & 5 \\
\midrule

I am a little confused about the test set performance on these GLUE datasets, are the labels to these sets not available to the public? & The comment raises a question about the test set performance on GLUE datasets and whether the labels are publicly available. However, it does not specify which part of the paper this question pertains to. Without explicit references, the authors cannot confidently determine which part of the paper needs clarification. The comment is not specific in terms of how to address the confusion. Therefore, this comment aligns with a score of 1. & \groundingshort & 5 & 2 \\
\midrule

I find the explanation in lines 243--245 very unclear. Could you please elaborate what this exactly means—why is it 5*3 (and not 5*2, e.g., in the case of beta-VAE where there's a mean and stdev in the code)? Also, line 247—why was k-NN used, and not some other more elaborate classifier? What is the value of k, and what is the distance metric used? & The comment consists of requests for clarification and questions about specific methodological choices in the paper. It does not contain opinions, judgments, or suggestions that require verification. The inquiries are factual and descriptive. & \verifiability & 3 & 2 \\
\midrule

The motivation of this paper is unclear. Existing research has widely recognized the importance of ensuring diversity in instruction-tuning data. However, this paper seems to oppose this common understanding without strong justification. The experiments do not persuade me, as they are somewhat weak: both the dataset and the LLM size are limited. The results are unconvincing and, if not thoroughly validated, could potentially mislead the community. & The comment identifies a critical issue with the paper's motivation, noting that it contradicts widely recognized research without providing strong justification. It also critiques the experiments for being weak due to limitations in both the dataset and the LLM size. While the comment highlights significant weaknesses, it lacks specific suggestions or guidance on how the authors could address these issues. & \helpfulness & 4 & 3 \\
\bottomrule
\end{tabular}
}
\caption{Samples from the manually evaluated review rationales, along with their scores for Relevance, and Correctness. The rationales have been generated by GPT-4o and are part of the synthetic \datasetname test set.}
\label{tab:rationale_examples}
\end{table*}

In Table~\ref{tab:rationale_examples} we show some samples for the manually annotated rationales, along with their scores.
\section{Aspect Score Rationale Evaluation}
Tables \ref{tab:rationale_actionability_grounding} and \ref{tab:rationale_verifiability_helpfulness} shows the full evaluation metrics for the rationale generation task,
\begin{table*}
\centering
\resizebox{1\textwidth}{!}{
\begin{tabular}{@{}lcccccccccccc@{}} 
\midrule
                  & \multicolumn{6}{c}{\textbf{\actionability}}                                       & \multicolumn{6}{c}{\textbf{\grounding}}                                 \\ 
\cmidrule(lr){2-7}
\cmidrule(lr){8-13}
\textbf{Model} & $\#C$ & $\#W\downarrow$  & $R_C$ & $R_W$ & $BS_C$ & $BS_W$ & $\#C$ & $\#W\downarrow$ & $R_C$ & $R_W$ & $BS_C$ & $BS_W$ \\
\midrule
\multicolumn{13}{c}{\textbf{Zero-Shot}}                                                           \\ \hdashline                                               
GPT-4o               & \textbf{736} & \textbf{246} & \textbf{0.355} & \textbf{0.328} & \textbf{0.720} & \textbf{0.694} & 584          & 398          & \textbf{0.362} & \textbf{0.278} & \textbf{0.703} & \textbf{0.672}  \\
Selene-1-Mini-8B     & 611          & 340          & 0.277          & 0.268          & 0.683          & \underline{0.659}  & \textbf{800} & \textbf{151} & \underline{0.288}  & \underline{0.272}  & \underline{0.672}  & \underline{0.657}   \\
Flow-Judge-v0.1-3.8B & 388          & 327          & \underline{0.297}  & \underline{0.271}          & \underline{0.684}  & 0.658          & 385          & 324          & 0.272          & 0.240           & 0.657          & 0.632           \\
Llama-3.1-8B         & 507          & 438          & 0.274          & 0.251          & 0.666          & 0.641          & 278          & 666          & 0.276          & 0.244          & 0.637          & 0.607           \\
Llama-3.1-IT-8B      & 502          & 312          & 0.219          & 0.210          & 0.634          & 0.610          & \underline{621}  & \underline{191}  & 0.247          & 0.222          & 0.630          & 0.626           \\
DeepSeek-R1-7B       & \underline{627}  & 367          & 0.210          & 0.208          & 0.622          & 0.609          & 595          & 388          & 0.213          & 0.186          & 0.615          & 0.592           \\
DeepSeek-R1-70B  & \underline{627} & 368 & 0.281 & 0.259 & 0.672 & 0.651 & 639 & 356 & 0.258 & 0.216 & 0.647 & 0.628  \\
Gemma-3-IT-27B   & 689 & 296 & 0.274 & 0.256 & 0.656 & 0.641 & 543 & 442 & 0.319 & 0.261 & 0.674 & 0.647  \\
SciTülu-7B           & 550          & \underline{291}  & 0.189          & 0.198          & 0.617          & 0.601          & 435          & 322          & 0.221          & 0.231          & 0.611          & 0.604           \\
SciLitLLM-7B         & 619          & 351          & 0.222          & 0.213          & 0.618          & 0.602          & 430          & 463          & 0.231          & 0.215          & 0.607          & 0.579           \\
DeepReviewer-7B      & 517          & 372          & 0.274          & 0.265          & 0.654          & 0.636          & 453          & 342          & 0.265          & 0.244          & 0.637          & 0.617           \\

\midrule
\multicolumn{13}{c}{\textbf{Fine-Tuning}}                  \\     \hdashline                           
Llama-3.1-8B         & \underline{792}  & 176          & \underline{0.502}  & 0.408          & \underline{0.784}  & 0.730          & \underline{743}  & 225          & \textbf{0.560} & 0.385          & \textbf{0.801} & \underline{0.732}   \\
Llama-3.1-IT-8B      & \textbf{802} & \underline{159}  & \textbf{0.503} & \textbf{0.418} & \textbf{0.785} & \underline{0.736}  & \textbf{758} & \underline{203}  & 0.556          & \textbf{0.389} & \underline{0.799}  & \textbf{0.736}  \\
DeepSeek-R1-7B       & 791          & \textbf{143} & 0.494          & \underline{0.412}  & 0.781          & \underline{0.736}  & 732          & \textbf{202} & 0.546          & 0.386          & 0.795          & \underline{0.732}   \\
SciTülu-7B           & 758          & 213          & 0.486          & 0.406          & 0.775          & 0.729          & 681          & 290          & \underline{0.557}  & 0.371          & \underline{0.799}          & 0.726           \\
SciLitLLM-7B         & 764          & 195          & 0.487          & 0.409          & 0.777          & 0.733          & 712          & 247          & 0.546          & \underline{0.387}  & 0.794          & 0.730           \\
DeepReviewer-7B      & 718          & 252          & 0.477          & 0.409          & 0.774          & \textbf{0.738} & 658          & 312          & 0.515          & 0.375          & 0.782          & 0.727           \\

\bottomrule
\end{tabular}
}

\caption{Rouge-L ($R$) and BERTScore ($BS$) metrics for the generated rationales compared with the rationales in the synthetic \datasetname test data. We split the evaluation by correctness of the predicted aspect score, considering predictions $\pm 1$ to the ground-truth label to be correct (indicated by subscript $C$), and predictions with a greater deviation to be incorrect (indicated by subscript $W$).}
\label{tab:rationale_actionability_grounding}
\end{table*}

\begin{table*}
\centering
\resizebox{0.95\textwidth}{!}{
\begin{tabular}{@{}lcccccccccccc@{}} 
\midrule
                  & \multicolumn{6}{c}{\textbf{\verifiability}}                                       & \multicolumn{6}{c}{\textbf{\helpfulness}}                    \\ 
\cmidrule(lr){2-7}
\cmidrule(lr){8-13}
\textbf{Model} & $\#C$ & $\#W\downarrow$ & $R_C$ & $R_W$ & $BS_C$ & $BS_W$ & $\#C$ & $\#W\downarrow$ & $R_C$ & $R_W$ & $BS_C$ & $BS_W$ \\
\midrule
\multicolumn{13}{c}{\textbf{Zero-Shot}}                                                           \\ \hdashline                                               
GPT-4o               & \underline{617}  & 365          & \textbf{0.319} & \textbf{0.284} & \textbf{0.685} & \textbf{0.652} & \textbf{928} & \underline{54}  & \textbf{0.264} & \underline{0.221}  & \textbf{0.674} & \underline{0.633}   \\
Selene-1-Mini-8B     & \textbf{697} & \underline{254}  & 0.275          & \underline{0.271}  & \underline{0.659}  & \underline{0.627}  & \underline{902}  & \textbf{49} & 0.231          & 0.215          & 0.652          & 0.628           \\
Flow-Judge-v0.1-3.8B & 222          & 487          & 0.250           & 0.241          & 0.649          & 0.617          & 495          & 215         & 0.238          & 0.213          & \underline{0.653}  & 0.626           \\
Llama-3.1-8B         & 272          & 672          & \underline{0.286}  & 0.236          & 0.654          & 0.612          & 276          & 668         & 0.237          & 0.214          & 0.643          & 0.619           \\
Llama-3.1-IT-8B      & 472          & 342          & 0.216          & 0.214          & 0.612          & 0.589          & 717          & 97          & 0.189          & 0.172          & 0.608          & 0.590           \\
DeepSeek-R1-7B       & 467          & 525          & 0.189          & 0.176          & 0.582          & 0.562          & 765          & 229         & 0.146          & 0.133          & 0.573          & 0.560           \\
DeepSeek-R1-70B   & 656 & 339 & 0.211 & 0.173 & 0.603 & 0.573 & 845 & 150 & 0.163 & 0.148 & 0.601 & 0.586  \\
Gemma-3-IT-27B   & 646 & 339 & 0.253 & 0.228 & 0.638 & 0.609 & 862 & 123 & 0.230  & 0.210  & 0.638 & 0.618  \\
SciTülu-7b           & 338          & \textbf{251} & 0.177          & 0.175          & 0.592          & 0.570          & 423          & 428         & 0.169          & 0.154          & 0.601          & 0.587           \\
SciLitLLM-7B         & 371          & 609          & 0.213          & 0.197          & 0.592          & 0.572          & 598          & 386         & 0.203          & 0.192          & 0.609          & 0.588           \\
DeepReviewer-7B      & 446          & 435          & 0.285          & 0.248          & 0.643          & 0.612          & 697          & 180         & \underline{0.246}  & \textbf{0.243} & 0.641          & \textbf{0.634}  \\

\midrule
\multicolumn{13}{c}{\textbf{Fine-Tuning}}                  \\     \hdashline                           
Llama-3.1-8B         & \textbf{758} & \underline{210}  & \underline{0.511}  & \textbf{0.369} & \textbf{0.773} & 0.689          & \underline{932}  & 36          & \underline{0.470}  & \underline{0.390}  & \underline{0.775}  & \underline{0.738}   \\
Llama-3.1-IT-8B      & \underline{741}  & 220          & \textbf{0.513} & 0.366          & \textbf{0.773} & 0.691          & 918          & 43          & \textbf{0.475} & 0.381          & \textbf{0.777} & 0.736           \\
DeepSeek-R1-7B       & 737          & \textbf{197} & 0.510          & 0.356          & \underline{0.772}  & 0.684          & 901          & \underline{33}  & 0.464          & 0.384          & 0.773          & 0.734           \\
SciTülu-7B           & 722          & 249          & 0.501          & 0.359          & 0.769          & 0.687          & \textbf{933} & 38          & 0.453          & 0.378          & 0.767          & 0.724           \\
SciLitLLM-7B         & 685          & 274          & 0.499          & 0.367          & 0.769          & \underline{0.692}  & \textbf{933} & \textbf{26} & 0.454          & 0.357          & 0.769          & 0.726           \\
DeepReviewer-7B      & 660          & 309          & 0.478          & \underline{0.368}  & 0.758          & \textbf{0.695} & 907          & 63          & 0.438          & \textbf{0.410} & 0.762          & \textbf{0.740}  \\

\bottomrule
\end{tabular}}

\caption{Rouge-L ($R$) and BERTScore ($BS$) metrics for the generated rationales compared with the rationales in the synthetic \datasetname test data. We split the evaluation by correctness of the predicted aspect score, considering predictions $\pm 1$ to the ground-truth label to be correct (indicated by subscript $C$), and predictions with a greater deviation to be incorrect (indicated by subscript $W$).}
\label{tab:rationale_verifiability_helpfulness}
\end{table*}

\section{Aspects Correlation}

\begin{figure}
    \centering
    \includegraphics[width=1\linewidth]{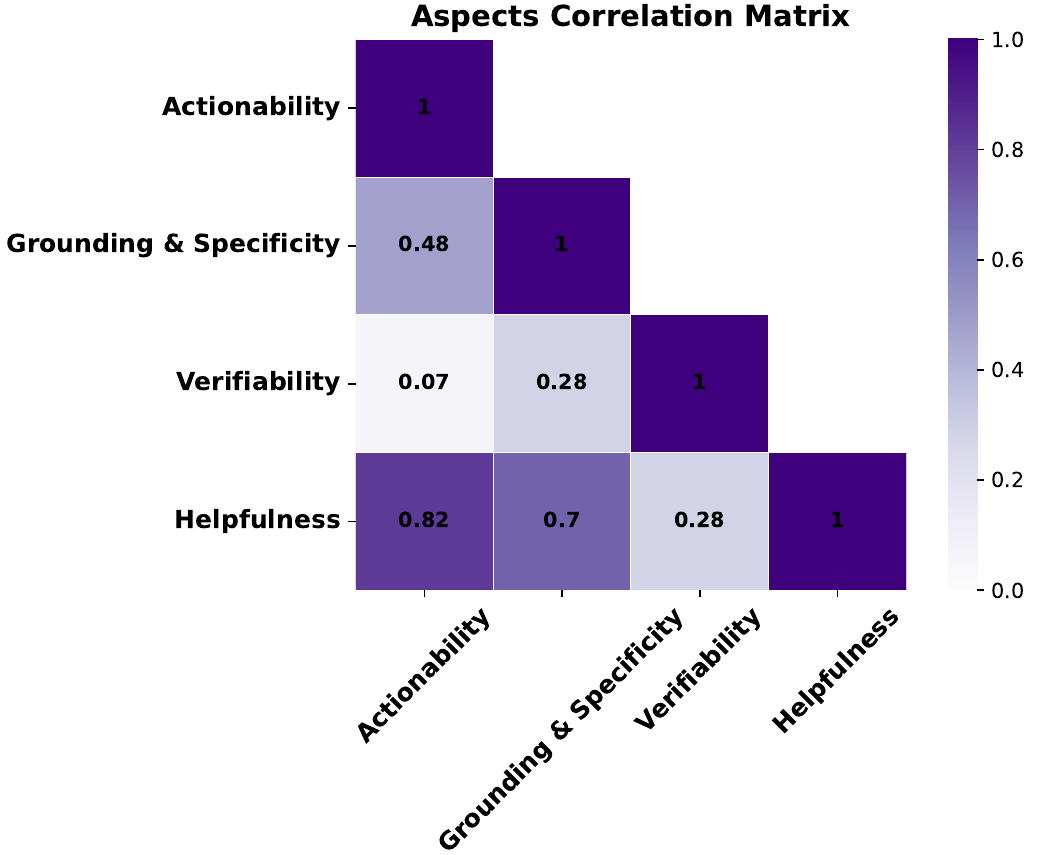}
    \caption{Pearson correlation matrix between the aspects' scores.}

    \label{fig:aspects_correlation}
\end{figure}

To examine the relationships between the different aspects, we compute the Pearson correlation coefficient for each pair of aspects using the majority vote of the human annotations. As shown in Fig.~\ref{fig:aspects_correlation}, \helpfulness exhibits the highest correlations with the other aspects, particularly with \actionability ($r = 0.82$) and \grounding ($r = 0.70$). This high correlation is expected, as the \helpfulness aspect acts as a collective measurement for the other aspects. In contrast, \verifiability shows relatively weak correlations with the other aspects, indicating it captures a more distinct dimension of review comment utility.

In Table~\ref{tab:data_samples} we show some samples from our \datasetname dataset. 

\begin{figure*}
    \centering
    \includegraphics[width=0.9\linewidth]{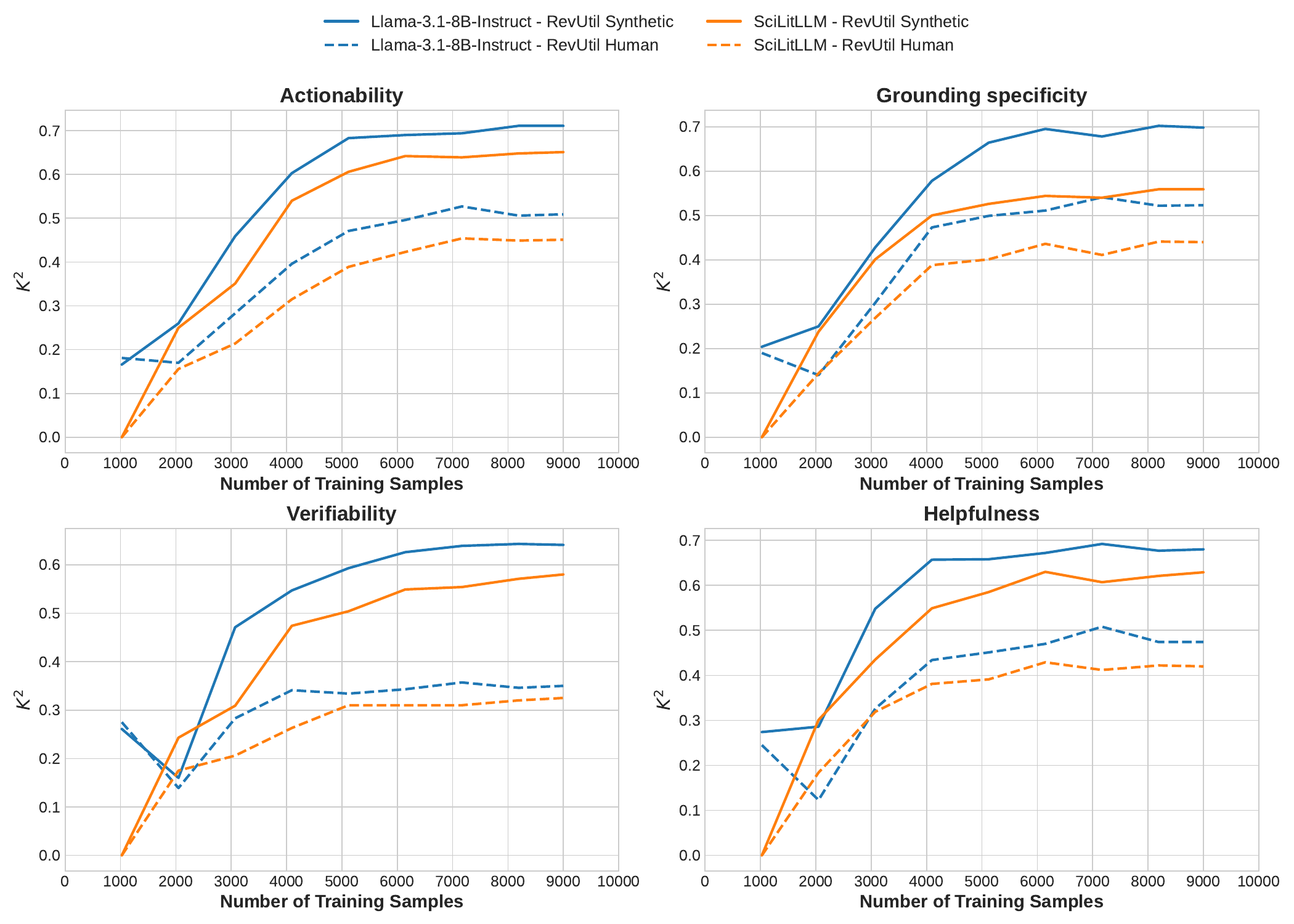}
    \caption{Model performance measured by $k^2$ across different percentages of training data.}

    \label{fig:progression_plot}
\end{figure*}

\section{Training Data Vs. Performance}
In this experiment, we measure the progression of the performance with the number of training samples. We chose two models: Llama-3.1-8B-Instruct and SciLitLLM. Results in Figure~\ref{fig:progression_plot} show that the models' performance starts to saturate after around 5k training examples. 

\begin{table}
\centering
\resizebox{\linewidth}{!}{
\begin{tabular}{@{}lcccc@{}} 
\toprule
\textbf{Context}                 &  \textbf{$\kappa^2$} & \textbf{$\rho$} & \textbf{$\alpha$} & F1   \\ 
\midrule
\multicolumn{5}{c}{\textbf{\actionability}}                                                                                        \\ 
\hdashline
Review Comment         & 0.357                    & 0.452                         & 0.388                      & -      \\
Paper + Review Comment & 0.245                    & 0.376                         & 0.337                      & -      \\ 
\midrule
\multicolumn{5}{c}{\textbf{\grounding}}                                                                                            \\ 
\hdashline
Review Comment         & 0.512                    & 0.510                          & 0.515                      & -      \\
Paper + Review Comment & 0.483                    & 0.465                         & 0.467                      & -      \\ 
\midrule
\multicolumn{5}{c}{\textbf{\verifiability}}                                                                                        \\ 
\hdashline
Review Comment        & 0.160                     & 0.460                          & 0.261                      & 0.368  \\
Paper + Review Comment & 0.230                     & 0.510                          & 0.288                      & 0.371  \\ 
\midrule
\multicolumn{5}{c}{\textbf{\helpfulness}}                                                                                        \\ 
\hdashline
Review Comment        & 0.318                    & 0.458                         & 0.352                      & -      \\
Paper + Review Comment & 0.224                    & 0.336                         & 0.302                      & -      \\
\bottomrule
\end{tabular}
}

\caption{Results for GPT-4o model on 100 samples under two settings: given the paper text as an extra context for the model, and without. }
\label{tab:paper_text_context_resutls}
\end{table}

\begin{table}[]
\centering

\begin{tabular}{@{}lc@{}}
\toprule
\textbf{Aspect}           & \textbf{Agree ($\pm1$)} \\ \midrule
\actionability         & 82\%                     \\
\groundingshort     & 93\%                     \\
\verifiability        & 89\%                     \\
\helpfulness           & 93\%                     \\
\midrule
AVG              & 89.25\%                  \\ \bottomrule
\end{tabular}
\caption{Agreement of score predictions by $\pm1$ of GPT-4o on 100 random samples when provided with and without the paper as context.}
\label{tab:paper_context_deviations}
\end{table}

\newpage
\section{Paper Text Context}
\label{app:paper_context_exp}
Our aspect definitions (\S \ref{ssec:aspect_definitions}) are designed to only require the review point context to fully judge the utility aspects, but to investigate whether providing extra context for the model would affect the model's performance, we randomly chose 100 review comments from the Reviewer2 dataset that we also have a majority human label for. We design two experiments, where in the first we give GPT-4o the full paper text (excluding the references), and in the second experiment, we do not provide the extra details, i.e., only the review comment. Results in Table~\ref{tab:paper_text_context_resutls} show that the extra context confused the model for all aspects except for \verifiability, where it leads to an increased performance.
We further analyze the predictions from our two settings, focusing on their agreement and deviation. Table~\ref{tab:paper_context_deviations} shows that in 89.25\% of samples, on average, the two setups arrive at the same prediction or their labels differ by at most ±1. This demonstrates that GPT-4o rarely uses the paper context to produce a significantly different prediction.

\end{document}